\documentclass[lettersize,journal]{IEEEtran}

\usepackage{amsmath,amsfonts,amssymb}
\usepackage{algorithmic}
\usepackage{algorithm}
\usepackage{array}
\usepackage[caption=false,font=normalsize,labelfont=sf,textfont=sf]{subfig}
\usepackage{textcomp}
\usepackage{stfloats}
\usepackage{url}
\usepackage{array}
\usepackage{makecell}
\usepackage{multirow}
\usepackage{verbatim}
\usepackage{graphicx}
\usepackage{wrapfig}
\usepackage{cite}
\usepackage{hyperref}
\usepackage{bm}
\usepackage{xfrac}
\usepackage{mathtools}
\usepackage{enumitem}
\hypersetup{colorlinks=true,
            linkcolor=magenta,
            urlcolor=magenta,
            citecolor=green}
\newcommand{\PreserveBackslash}[1]{\let\temp=\\#1\let\\=\temp}
\newcolumntype{C}[1]{>{\PreserveBackslash\centering}p{#1}}
\newcolumntype{R}[1]{>{\PreserveBackslash\raggedleft}p{#1}}
\newcolumntype{L}[1]{>{\PreserveBackslash\raggedright}p{#1}}

\newcommand{\zsyrevise}[1]{\textcolor{black}{#1}}

\DeclareMathOperator\dif{d\!}

\begin{document}

\title{Efficient Diffusion Model for Image Restoration by Residual Shifting}

\author{Zongsheng Yue,~Jianyi Wang,~Chen Change Loy,~\IEEEmembership{Senior Member,~IEEE}
\thanks{Z. Yue, J. Wang, and C. C. Loy are with S-Lab, Nanyang Technological University (NTU), Singapore (E-mail: zsyue@gmail.com, \{jianyi001,ccloy\}@ntu.edu.sg).}
\thanks{C. C. Loy is the corresponding author.}
}

\markboth{}%
{Shell \MakeLowercase{\textit{et al.}}: A Sample Article Using IEEEtran.cls for IEEE Journals}


\maketitle

\begin{abstract}
While diffusion-based image restoration (IR) methods have achieved remarkable success, they are still limited by the low inference speed attributed to the necessity of executing hundreds or even thousands of sampling steps. Existing acceleration sampling techniques, though seeking to expedite the process, inevitably sacrifice performance to some extent, resulting in over-blurry restored outcomes. To address this issue, this study proposes a novel and efficient diffusion model for IR that significantly reduces the required number of diffusion steps. Our method avoids the need for post-acceleration during inference, thereby avoiding the associated performance deterioration. Specifically, our proposed method establishes a Markov chain that facilitates the transitions between the high-quality and low-quality images by shifting their residuals, substantially improving the transition efficiency. A carefully formulated noise schedule is devised to flexibly control the shifting speed and the noise strength during the diffusion process. Extensive experimental evaluations demonstrate that the proposed method achieves superior or comparable performance to current state-of-the-art methods on \zsyrevise{four classical IR tasks, namely image super-resolution, image inpainting, blind face restoration, and image deblurring,} \textit{\textbf{even only with four sampling steps}}. Our code and model are publicly available at \url{https://github.com/zsyOAOA/ResShift}.
\end{abstract}


\section{Introduction}
\IEEEPARstart{I}mage restoration (IR) \zsyrevise{is a critical challenge in the field of low-level vision, with the goal of recovering a high-quality (HQ) image from its corresponding low-quality (LQ) variant. This challenge can be further divided into different sub-tasks upon its degradation model, including image super-resolution~\cite{dong2012nonlocally,dong2015image}, image deblurring~\cite{kupyn2019deblur,whang2022deblurring}, and image inpainting~\cite{yu2019free,suvorov2022resolution}, among others~\cite{gu2017weighted,zhang2017beyond}. Particularly, the degradation models encountered in practical scenarios, such as those in real-world super-resolution, often exhibit significant complexity, rendering the IR problem severely ill-posed and challenging to address.}

\zsyrevise{The diffusion model~\cite{sohl2015deep} has revolutionized the traditional paradigm of image generation based on Generative Adversarial Networks (GANs)~\cite{goodfellow2014generative,brock2018large}, further advancing the field of image synthesis~\cite{ho2020denoising,dhariwal2021diffusion}. This approach leverages a hidden Markov chain to progressively corrupt an image into white Gaussian noise through a forward diffusion process, and subsequently employs a deep neural network to approximate the reverse process for image reconstruction. Attributed to its powerful generative capability, the diffusion model has shown considerable potential in addressing various IR tasks, including image denoising~\cite{delbracio2023inversion,luo2023image}, deblurring~\cite{whang2022deblurring,murata23agibbs}, inpainting~\cite{meng2021sdedit,Lugmayr2022repaint,chung2022come,yue2022difface}, colorization~\cite{song2021scorebased,saharia2022palette,liang2024control}. The exploration of diffusion models' capabilities in IR still remains an active and promising area of research.}

\begin{figure*}[t]
    \centering
    \includegraphics[width=\linewidth]{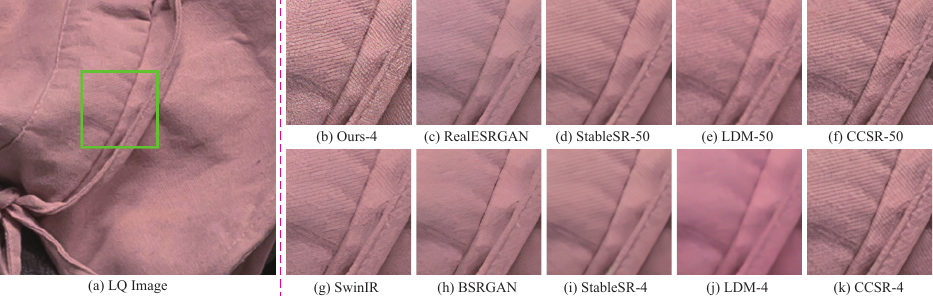}
    \vspace{-6mm}
    \caption{\zsyrevise{Qualitative comparisons on one typical real-world example of the proposed method and recent state-of-the-arts, including RealESRGAN~\cite{wang2021real}, BSRGAN~\cite{zhang2021designing}, SwinIR~\cite{liang2021swinir}, LDM~\cite{rombach2022high}, StableSR~\cite{wang2023exploiting}, and CCSR~\cite{sun2023improving}. As for the diffusion-based approaches and our proposed method, we annotate the number of sampling steps with the format of ``Method-A'' for more intuitive visualization, where ``A'' denotes the number of sampling steps. 
    }}
    \label{fig:compare_ldm} \vspace{-2mm}
\end{figure*}

\zsyrevise{In this study, we categorize recent diffusion-based IR methods into two main approaches. The first approach~\cite{saharia2022image,saharia2022palette,rombach2022high,li2022srdiff,Xia_2023_ICCV} directly incorporates the LQ image into the input of a current diffusion model, such as DDPM~\cite{ho2020denoising}, as a condition, and then retrain this model specifically for the IR task. Once trained, this model can generate the desirable HQ image from Gaussian noise and the observed LQ image through the reverse sampling process.
The second approach, as explored in~\cite{kawar2022denoising,chung2022improving,wang2023exploiting,yang2023pixel,sun2023coser,wang2023zeroshot,zhu2023denoising,wu2023seesr}, adopts a pre-trained unconditional diffusion model as a prior to address the IR problem. This method modifies the reverse sampling procedure to align the generated outputs with the given LQ observations by incorporating the degradation model at each iteration. However, both strategies are limited by the inherent Markov chain structure of DDPM, which often leads to inefficiencies during inference, requiring hundreds or even thousands of sampling steps.
While recent advancements~\cite{nichol2021improved,song2021denoising,lu2022dpmsolver} have introduced acceleration techniques to reduce sampling steps, these methods inevitably result in a significant performance drop, as evidenced by the overly smooth results shown in Fig.~\ref{fig:compare_ldm}(i)-(k). Thus, there is a need to devise a new diffusion model specifically designed for IR that achieves an optimal trade-off between efficiency and performance, without sacrificing one for the other.}

\zsyrevise{In the domain of image generation, diffusion models progressively convert the observed data into a pre-determined prior distribution, often a standard Gaussian distribution, through a Markov chain over numerous steps. The forward diffusion process constructs this Markov chain, while the reverse process involves training a deep neural network to approximate the inverse trajectory of the Markov chain. The trained neural network can generate images randomly by sampling from the reverse Markov chain, initiating at the Gaussian distribution. Although the Gaussian distribution is well-suited for image generation, its optimality is questioned for IR, where LQ images are available as extra information. 
In this paper, we argue that a reasonable diffusion model for IR should start from a prior distribution centered around the LQ image, enabling an iterative recovery of the HQ image from its LQ counterpart instead of Gaussian white noise. This approach not only aligns more closely with the characteristics of IR but also holds the potential to reduce the number of diffusion steps for sampling, thereby improving inference efficiency.}

\zsyrevise{In light of the preceding motivation, we introduce an efficient diffusion model characterized by a shorter Markov chain transferring between the HQ and LQ images. The Markov chain's initial state converges towards an approximate distribution of the HQ image, while the final state approximates the LQ image distribution. This is achieved through the design of a transition kernel that incrementally shifts residual information between the HQ and LQ image pair. 
Our method exhibits superior efficiency beyond existing diffusion-based IR methods, due to its capacity to rapidly transfer residual information across a limited number of steps. Moreover, this design also allows for an analytical and concise expression of the evidence lower bound (ELBO), thereby simplifying the formulation of the optimization objective for training. Beyond the traditional ELBO, we empirically find that introducing a perceptual regularizer can further reduce the diffusion steps during training,  and thus improve the inference efficiency.  
Building upon the constructed diffusion kernel, we develop a highly flexible noise schedule that controls the rate of residual transfer and the intensity of the added noise at each step. This schedule provides a mechanism for balancing the fidelity and realism of the recovered images by tuning its hyper-parameters.}

In summary, the main contributions of this work are as follows:
\begin{itemize}[topsep=0pt,parsep=0pt,leftmargin=18pt]
    \item We propose an efficient diffusion model specifically for IR.
    It builds up a short Markov chain between the HQ/LQ images, rendering a fast reverse sampling trajectory during inference. Extensive experiments show that our approach \textbf{\textit{requires only four sampling steps}} to achieve appealing results, outperforming or at least being comparable to current state-of-the-art methods. A preview of the comparison results of the proposed method to recent approaches is shown in Fig.~\ref{fig:compare_ldm}.
    \item A highly flexible noise schedule is designed for the proposed diffusion model, capable of controlling the transition properties more precisely, including the shifting speed and the noise level. Through tuning the hyper-parameters, our method offers a more graceful solution to the widely acknowledged perception-distortion trade-off in IR.
    \item \zsyrevise{Based on the traditional diffusion Unet, we propose to substitute its self-attention layers with Swin Transformer blocks to enhance its capability in handling images with varying resolutions.}
    \item The proposed method is a general diffusion-based framework for IR and capable of handling various IR tasks. \zsyrevise{This study has thoroughly substantiated its effectiveness and superiority on four typical and challenging IR tasks, namely image super-resolution, image inpainting, blind face restoration, and image deblurring.} 
\end{itemize}

In summary, our work formulates an efficient diffusion model tailored for IR, overcoming the limitation of prevailing approaches on inference efficiency. A preliminary version of this work has been published in NeurIPS 2023~\cite{yue2023resshift}, focusing only on the task of image super-resolution. This study makes substantial improvements in both model design and empirical evaluation across diverse IR tasks compared with the conference version. Concretely, we incorporate the perceptual loss into the model optimization and substitute the self-attention layer with shifted window-based self-attention presented in Swin Transformer~\cite{liu2021swin} in the network architecture. The former modification can further reduce the diffusion steps from 15 to 4, and the latter endows our model with graceful adaptability to handle arbitrary resolutions during inference.

The remainder of the manuscript is organized as follows: Section~\ref{sec:related-work} introduces the related work. Section~\ref{sec:method} presents our designed diffusion model for IR. In Section~\ref{sec:exp-sr} and Section~\ref{sec:exp-inpainting}, extensive experiments are conducted to evaluate the performance of our proposed method on the task of image super-resolution and image inpainting, respectively. Section~\ref{sec:conclusion} finally concludes the paper.

\section{Related Work} \label{sec:related-work}
In this section, we briefly review the literature on image restoration, traversing from conventional non-diffusion methodologies to recent diffusion-based approaches.
\subsection{Conventional Image Restoration Approaches}
Most of the conventional IR methods can be cast into the Maximum a Posteriori (MAP) framework, a Bayesian paradigm encompassing a likelihood (loss) term and a prior (regularization) term. The likelihood reflects the underlying noise distribution of the LQ image. The commonly used $L_2$ or $L_1$ loss indeed corresponds to a Gaussian or Laplacian assumption on image noise. To more accurately depict the noise distribution, some robust alternatives were introduced, such as Poissonian-Gaussian~\cite{foi2008practical}, MoG~\cite{meng2013robust}, MoEP~\cite{cao2016robust}, Dirichlet MoG~\cite{zhu2016blind,yue2018hyperspectral} and so on. 
%
Simultaneously, there has been an increased focus on employing image priors to address the inherent ill-posedness of IR over recent decades.
Typical image priors encompass total variation~\cite{rudin1992nonlinear}, wavelet coring~\cite{simoncelli1996noise}, non-local similarity~\cite{buades2005non,dong2012nonlocally}, sparsity~\cite{dong2011image,gu2015convolutional}, low-rankness~\cite{gu2017weighted,xu2017multichannel}, dark channel~\cite{he2010single,pan2016blind}, among others. These conventional methods are mainly limited by the model capacity and the subjectivity inherited from the manually designed assumptions on image noise and prior. 

In recent years, the landscape of IR has been dominated by deep learning (DL)-based methodologies. The seminal works~\cite{dong2015image,sun2015learning,zhang2017beyond} proposed to solve the IR problem using a convolution neural network, outperforming traditional model-based methods on the tasks of image denoising, super-resolution, and deblurring. Then, many studies~\cite{shi2016real,tai2017memnet,lai2017deep,haris2018deep,zhang2018residual,hu2019meta,wan2021high,zamir2022restormer,wang2022uformer,suvorov2022resolution} have emerged, mainly concentrating on designing more delicate network architectures to further improve the restoration performance. Besides, there have been some discernible investigations that seek to combine current DL tools and classical IR modeling ideas. Notable works include the plug-and play or unfolding paradigm~\cite{zhang2018learning,simon2019rethinking,zhang2020deep}, learnable image priors~\cite{liu2018non,ulyanov2018deep,gu2020multigan,pan2022exploitingdgp}, and the loss-oriented methods~\cite{wang2018esrgan,yue2022blind,lugmayr2020srflow}. The infusion of deep neural networks, owing to their large model capacity, has substantively extended the frontiers of IR tasks.

\subsection{Diffusion-based Image Restoration Approaches}
Inspired by principles from non-equilibrium statistical physics, Sohl-Dickstein \MakeLowercase{\textit{et al.}}~\cite{sohl2015deep} proposed the diffusion model to fit complex distributions. Subsequent advancements by Ho \MakeLowercase{\textit{et al.}}~\cite{ho2020denoising} and Song \MakeLowercase{\textit{et al.}}~\cite{song2021scorebased} further improve its theoretical foundation by integrating denoising score matching and stochastic differential equation, thereby achieving impressive success in image generation~\cite{rombach2022high,parmar2024one}. Owing to its powerful generative capability, diffusion models have also found successful applications in the field of IR. Next, we provide a comprehensive overview of recent developments in diffusion-based IR methods.

The most straightforward solution to solve the IR problem using diffusion models is to introduce the LQ image as an addition condition in each timestep. Pioneering this approach, Saharia et al.~\cite{saharia2022image} have successfully trained a diffusion model for image super-resolution. Subsequent studies~\cite{saharia2022palette,whang2022deblurring,Xia_2023_ICCV} further expanded upon this approach, exploring its applicability in image deblurring, colorization, and denoising. To circumvent the resource-intensive process of training from scratch, 
an alternative strategy involves harnessing a pre-trained diffusion model to facilitate IR tasks. Numerous investigations, such as~\cite{chung2022come,chung2022improving,kawar2022denoising,Lugmayr2022repaint,wang2023zeroshot,zhu2023denoising,murata23agibbs,xiao2024dreamclean}, reformulated the reverse sampling procedure of a pre-trained diffusion model into an optimization problem by incorporating the degradation model, enabling solving the IR problem in a zero-shot manner. Most of these methods, however, cannot deal with the blind IR problem, as they rely on a pre-defined degradation model. In contrast, some other works~\cite{wang2023exploiting,zhang2023adding,yang2023pixel,lin2023diffbir,zhang2024unified} directly introduced a trainable module that takes the LQ image as input. This module modulates the feature maps of the pre-trained diffusion model, steering it toward the direction of generating a desirable HQ image. Such a paradigm eliminates the reliance on a degradation model in the test phase, rendering it capable of handling the blind IR tasks. 

The methodologies mentioned above are grounded in the foundational diffusion model initially crafted for image generation, necessitating a large number of sampling steps. This inefficiency presents a constraint on their application in real scenarios. The primary goal of our investigation is to devise a new diffusion model customized for IR, which facilitates a swift transition between the LQ/HQ image pair, thereby enhancing efficiency during inference.

\section{Methodology}\label{sec:method}
In this section, we present our proposed diffusion model tailored for IR. For ease of presentation, the LQ image and the corresponding HQ image are denoted as $\bm{y}_0$ and $\bm{x}_0$, respectively. Notably, we further assume $\bm{y}_0$ and $\bm{x}_0$ have identical spatial resolution, which can be easily achieved by pre-upsampling the LQ image $\bm{y}_0$ using nearest neighbor interpolation if necessary.

\begin{figure*}[t]
    \centering
    \includegraphics[width=\linewidth]{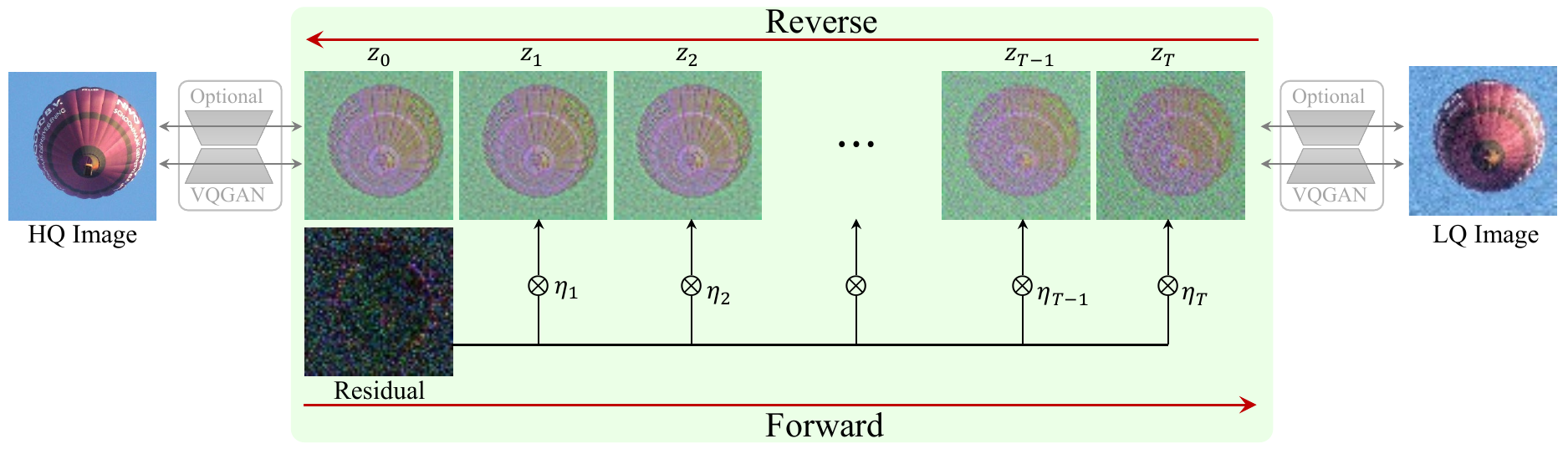}
    \vspace{-6mm}
    \caption{Overview of the proposed method. Our method builds up a Markov chain between the HQ/LQ image pair by shifting their residuals. To alleviate the computational burden of this transition, it can be optionally moved to the latent space of VQGAN~\cite{esser2021taming}.}
    \label{fig:framework}
    \vspace{-2mm}
\end{figure*}

\subsection{Model Design}
The iterative sampling paradigm of diffusion models has proven highly effective in generating intricate and vivid image details, inspiring us to design an iterative approach to address the IR problem as well. Our proposed method builds up a Markov chain to facilitate a transition from the HQ image to its LQ counterpart as shown in Fig.~\ref{fig:framework}. In this way, the restoration of the desirable HQ image can be achieved by sampling along the reverse trajectory of this Markov chain that starts at the given LQ image. Next, we will detail how to construct such a Markov chain specifically for IR.

\vspace{2mm}\noindent\textbf{Forward Process}. Let's denote the residual between the LQ image and its corresponding HQ counterpart as $\bm{e}_0$, i.e., $\bm{e}_0=\bm{y}_0-\bm{x}_0$. Our core idea is to construct a transition from $\bm{x}_0$ to $\bm{y}_0$ by gradually shifting their residual $\bm{e}_0$ through a Markov chain with length $T$. Before that, we first introduce a shifting sequence $\{\eta_t\}_{t=1}^T$, which increases monotonically with respect to timestep $t$ and adheres to the condition of $\eta_1 \to 0$ and $\eta_T \to 1$. Then, the transition distribution is formulated based on this shifting sequence as follows:
\begin{equation}
    q(\bm{x}_t|\bm{x}_{t-1},\bm{y}_0) = \mathcal{N}(\bm{x}_t; \bm{x}_{t-1}+\alpha_t \bm{e}_0, \kappa^2 \alpha_t \bm{I}), ~ t=1,\cdots,T,
    \label{eq:transit_t_t1}
\end{equation}
where $\alpha_1=\eta_1$ and $\alpha_t=\eta_t-\eta_{t-1}$ for $t>1$, $\kappa$ is a hyper-parameter controlling the noise variance, $\bm{I}$ is the identity matrix. Notably, we show that the marginal distribution at any timestep $t$ is analytically integrable, namely
\begin{equation}
    q(\bm{x}_t|\bm{x}_0, \bm{y}_0) = \mathcal{N}(\bm{x}_t; \bm{x}_0+\eta_t \bm{e}_0, \kappa^2 \eta_t \bm{I}), ~ t=1,\cdots,T.
    \label{eq:transit_0_t}
\end{equation}

The design of the transition distribution presented in Eq.~\eqref{eq:transit_t_t1} is guided by two primary principles. The first principle concerns the standard deviation, i.e., $\kappa\sqrt{\alpha_t}$, which aims to facilitate a smooth transition between $\bm{x}_t$ and $\bm{x}_{t-1}$. This is achieved by bounding the expected distance between $\bm{x}_t$ and $\bm{x}_{t-1}$ with $\sqrt{\alpha_t}$, given that the image data falls within the range of $[0, 1]$. Mathematically, this is expressed as:
\begin{equation}
    \text{max}[(\bm{x}_0+\eta_t\bm{e}_0)-(\bm{x}_0+\eta_{t-1}\bm{e}_0)] = \text{max}[\alpha_t\bm{e}_0]<\alpha_t<\sqrt{\alpha_t},
    \label{eq:var_bound}
\end{equation}
where $\text{max}[\cdot]$ represents the pixel-wise maximizing operation. The hyper-parameter $\kappa$ is introduced to increase the flexibility of this design.
The second principle pertains to the mean parameter, i.e., $\bm{x}_0+\alpha_t\bm{e}_0$. Combining with the definition of $\alpha_t$, namely $\alpha_t=\eta_t-\eta_{t-1}$, it induces the marginal distribution in Eq.~\eqref{eq:transit_0_t}. Furthermore, the marginal distributions of $\bm{x}_1$ and $\bm{x}_T$ converges to $\delta_{\bm{x}_0}(\bm{x})$\footnote{$\delta_{\bm{\mu}}(\bm{x})$ denotes the Dirac distribution centered at $\bm{\mu}$.} and $\mathcal{N}(\bm{x};\bm{y}_0, \kappa^2\bm{I})$, respectively, serving as approximations for the HQ image and the LQ image. By constructing the Markov chain in such a thoughtful way, it is possible to handle the IR task by inversely sampling from it given the LQ image $\bm{y}_0$.

\begin{algorithm}[t]
    \caption{\zsyrevise{Training}}
    \label{alg:training}
    \begin{algorithmic}[1]
        \REQUIRE \zsyrevise{Degradation model $\mathcal{D}(\cdot)$, high-quality dataset $\mathcal{T}$}
        \REPEAT
        \STATE \zsyrevise{$\bm{x}_0 \sim \mathcal{T}$, $\bm{y}_0 = \mathcal{D}(\bm{x}_0)$}
        \STATE \zsyrevise{$t \sim \text{Uniform}\left(\{1,\cdots,T\}\right)$}
        \STATE \zsyrevise{$\bm{x}_t \sim q(\bm{x}_t | \bm{x}_0, \bm{y}_0)$}
        \STATE \zsyrevise{Take gradient descent step on $\nabla \mathcal{L}_{\bm{\theta}}(\bm{x}_t, \bm{y}_0, t)$}
        \UNTIL{\zsyrevise{converged}}
    \end{algorithmic}
\end{algorithm}

\vspace{2mm}\noindent\textbf{Reverse Process}. The reverse process endeavors to estimate the posterior distribution $p(\bm{x}_0|\bm{y}_0)$ through the following formalization:
\begin{equation}
    p(\bm{x}_0|\bm{y}_0)=\int p(\bm{x}_T|\bm{y}_0)\prod_{t=1}^T p_{\bm{\theta}}(\bm{x}_{t-1}|\bm{x}_t,\bm{y}_0) \mathrm{d}\bm{x}_{1:T},
    \label{eq:poster_inverse}
\end{equation}
where $p(\bm{x}_T|\bm{y}_0) \approx \mathcal{N}(\bm{x}_T|\bm{y}_0, \kappa^2\bm{I})$, $p_{\bm{\theta}}(\bm{x}_{t-1}|\bm{x}_t,\bm{y}_0)$ represents the desirable inverse transition kernel from $\bm{x}_t$ to $\bm{x}_{t-1}$, parameterized by a learnable parameter $\bm{\theta}$. Consistent with prevalent literature of diffusion model~\cite{sohl2015deep,ho2020denoising,song2021scorebased}, we adopt the following Gaussian assumption: 
\begin{equation}
    \label{eq:format_posterior_gaussian}
   p_{\bm{\theta}}(\bm{x}_{t-1}|\bm{x}_t,\bm{y}_0)=\mathcal{N}\left(\bm{x}_{t-1};\bm{\mu}_{\bm{\theta}}(\bm{x}_t, \bm{y}_0, t), \bm{\Sigma}_{\bm{\theta}}(\bm{x}_t, \bm{y}_0, t)\right).  
\end{equation}
The optimization for $\bm{\theta}$ is achieved by minimizing the following negative ELBO, i.e., 
\begin{equation}
    \sum_t D_{\text{KL}}\left[q(\bm{x}_{t-1}|\bm{x}_t,\bm{x}_0, \bm{y}_0)\Vert p_{\bm{\theta}}(\bm{x}_{t-1}|\bm{x}_t,\bm{y}_0)\right],
    \label{eq:elbo}
\end{equation}
where $D_{\text{KL}}[\cdot\Vert\cdot]$ denotes the Kullback-Leibler (KL) divergence. More mathematical details can be found in~\cite{sohl2015deep} or \cite{ho2020denoising}.

\begin{algorithm}[t]
    \caption{\zsyrevise{Sampling}}
    \label{alg:sampling}
    \begin{algorithmic}[1]
        \REQUIRE \zsyrevise{Low-quality image $\bm{y}$}
        \STATE \zsyrevise{$\bm{x}_T \sim \mathcal{N}(\bm{x}_T;\bm{y}, \kappa^2 \eta_T \bm{I})$}
        \FOR{$t=T,\cdots,1$}
        \STATE \zsyrevise{$\bm{\epsilon} \sim \mathcal{N}(\bm{\epsilon};\bm{0},\bm{I})$ if $t>1$ else $\bm{\epsilon}=\bm{0}$}
        \STATE \zsyrevise{$\bm{\mu} = \frac{\eta_{t-1}}{\eta_t}\bm{x}_t+\frac{\alpha_t}{\eta_t}f_{\bm{\theta}}(\bm{x}_t,\bm{y},t)$}
        \STATE \zsyrevise{$\bm{x}_{t-1} = \bm{\mu} + \kappa \sqrt{\frac{\eta_{t-1}\alpha_t}{\eta_t}}  \bm{\epsilon}$}
        \ENDFOR
        \RETURN \zsyrevise{$\bm{x}_0$}
    \end{algorithmic}
\end{algorithm}
By combining Eq.~\eqref{eq:transit_t_t1} and Eq.~\eqref{eq:transit_0_t}, the target distribution $q(\bm{x}_{t-1}|\bm{x}_t,\bm{x}_0,\bm{y}_0)$ in Eq.~\eqref{eq:elbo} can be rendered tractable and expressed in an explicit form given below:
\begin{equation}
    q(\bm{x}_{t-1}|\bm{x}_t,\bm{x}_0,\bm{y}_0)=\mathcal{N}\left(\bm{x}_{t-1}\bigg\vert\frac{\eta_{t-1}}{\eta_t}\bm{x}_t+\frac{\alpha_t}{\eta_t}\bm{x}_0, 
    \kappa^2\frac{\eta_{t-1}}{\eta_t}\alpha_t\bm{I}\right).
    \label{eq:poster_elbo}
\end{equation}
The detailed calculation of this derivation can be found in Appendix~\ref{subsec:math_supp}. Considering that the variance parameter is independent of $\bm{x}_t$ and $\bm{y}_0$, we thus set it to be the true variance, i.e.,
\begin{equation}
    \label{eq:posterior_variance}
    \bm{\Sigma}_{\bm{\theta}}(\bm{x}_t,\bm{y}_0,t)=\kappa^2\frac{\eta_{t-1}}{\eta_t}\alpha_t\bm{I}.   
\end{equation}
The mean parameter $\bm{\mu}_{\bm{\theta}}(\bm{x}_t,\bm{y}_0,t)$ is reparameterized as:
\begin{equation}
    \bm{\mu}_{\bm{\theta}}(\bm{x}_t,\bm{y}_0,t) = \frac{\eta_{t-1}}{\eta_t}\bm{x}_t+\frac{\alpha_t}{\eta_t}f_{\bm{\theta}}(\bm{x}_t,\bm{y}_0,t),
    \label{eq:repqrameter_mean_reparameter}
\end{equation}
where $f_{\bm{\theta}}(\cdot)$ is a deep neural network with parameter $\bm{\theta}$, aiming to predict $\bm{x}_0$.  We explored different parameterization forms on $\bm{\mu}_{\theta}$ and found that  Eq.~\eqref{eq:repqrameter_mean_reparameter} exhibits superior stability and performance. 

Based on Eq.~\eqref{eq:repqrameter_mean_reparameter}, the objective function in Eq.~\eqref{eq:elbo} is then simplified as:
\begin{equation}
    \mathcal{L}_{\bm{\theta}}(\bm{x}_t, \bm{y}_0, t) = \sum\nolimits_t w_t \Vert \hat{\bm{x}}_0^t - \bm{x}_0 \Vert_2^2,
    \label{eq:loss_l2}
\end{equation}
where $w_t = \frac{\alpha_t}{2\kappa^2\eta_t\eta_{t-1}}$, $\hat{\bm{x}}_0^t=f_{\bm{\theta}}(\bm{x}_t, \bm{y}_0, t)$. In practice, we empirically find that the omission of weight $w_t$ results in a notable performance improvement, aligning with the conclusion in~\cite{ho2020denoising}. \zsyrevise{And the detailed training process is summarized in Algorithm~\ref{alg:training}. After training, we can generate the desirable HQ result following Algorithm~\ref{alg:sampling} for any LQ testing image.}

\begin{figure*}[t]
    \centering
    \includegraphics[width=\linewidth]{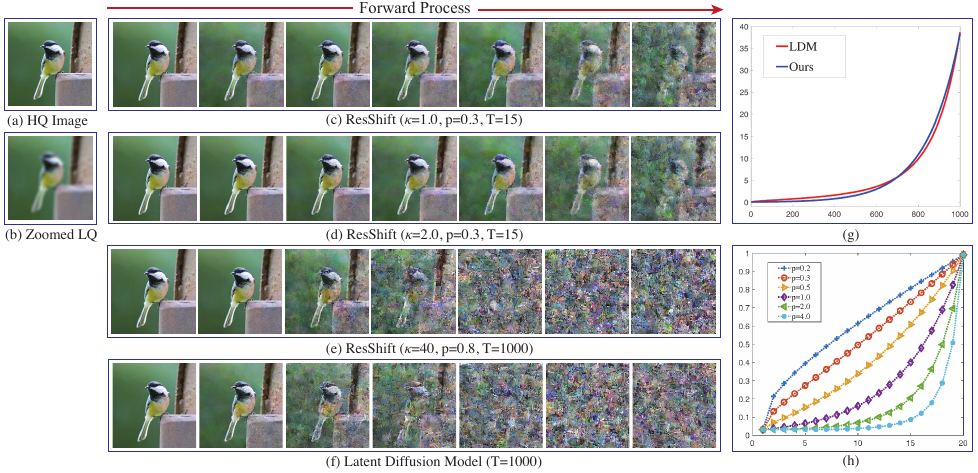}
    \vspace{-6mm}
    \caption{\zsyrevise{Illustration of the proposed noise schedule. (a) HQ image. (b) Zoomed LQ image. (c)-(d) Diffused images of the proposed noise schedule in timesteps of 1, 3, 5, 7, 9, 12, and 15 under different values of $\kappa$ by fixing $p=0.3$ and $T=15$. (e)-(f) Diffused images of our method with a specified configuration of $\kappa=40, p=0.8, T=1000$ and LDM~\cite{rombach2022high} in timesteps of 100, 200, 400, 600, 800, 900, and 1000. (g) The relative noise intensity (vertical axes, measured by $\sqrt{\sfrac{1}{\lambda_{\text{snr}}}}$, where $\lambda_{\text{snr}}$ denotes the signal-to-noise ratio) of the schedules in (d) and (e) w.r.t. the timesteps (horizontal axes). (h) The shifting speed $\sqrt{\eta_t}$ (vertical axes) w.r.t. to the timesteps (horizontal axes) across various configurations of $p$. Note that the diffusion processes in this figure are implemented in the latent space, but we display the intermediate results after decoding back to the image space for the purpose of easy visualization.}}
    \label{fig:schedule}
\end{figure*}
\vspace{2mm}\noindent\textbf{Perceptual Regularization}. As presented above, our proposed method facilitates an iterative restoration process starting from the LQ image, in contrast to prior methods that initialize from Gaussian noise. This approach effectively reduces the number of diffusion steps. The comprehensive experimental analysis in our conference paper~\cite{yue2023resshift} has substantiated that the proposed method yields promising results with a mere 15 sampling steps, demonstrating a notable acceleration compared to established methodologies~\cite{rombach2022high,wang2023exploiting}.

Unfortunately, attempts at further acceleration, particularly with fewer than 5 sampling steps, tend to produce over-smooth results. This phenomenon is primarily attributed to that the $L_2$-based loss in Eq.~\eqref{eq:loss_l2} favors the prediction of an average over plausible solutions. To overcome this limitation, we introduce an additional perceptual regularization~\cite{zhang2018unreasonable} on Eq.~\eqref{eq:loss_l2} to further constrain the solution space, namely,
\begin{equation}
    \mathcal{L}_{\bm{\theta}}(\bm{x}_t, \bm{y}_0, t) = \sum\nolimits_t \Vert \hat{\bm{x}}_0^t - \bm{x}_0 \Vert_2^2 + \lambda l_p\left(\hat{\bm{x}}_0^t, \bm{x}_0\right) ,
    \label{eq:loss_l2_lpips}
\end{equation}
where $l_p(\cdot, \cdot)$ denotes the pre-trained LPIPS metric, $\lambda$ is a hyper-parameter controlling the relative importance of these two constraints. This solution effectively curtails the sampling trajectory to fewer steps, e.g., 4 steps in this study, while concurrently maintaining superior performance.

\vspace{2mm}\noindent\textbf{Extension to Latent Space}. To alleviate the computational overhead in training, our proposed model can be optionally moved into the latent space of VQGAN~\cite{esser2021taming}, where the original image is compressed by a factor of four in spatial dimensions. This does not require any modifications on our model other than substituting $\bm{x_0}$ and $\bm{y}_0$ with their latent codes. An intuitive illustration is shown in Fig.~\ref{fig:framework}.

\subsection{Noise Schedule}
The proposed method employs a hyper-parameter $\kappa$ and a shifting sequence $\{\eta_t\}_{t=1}^{T}$ to determine the noise schedule of the diffusion process. In particular, the hyper-parameter $\kappa$ regulates the overall noise intensity during the transition, and its influence on performance is empirically discussed in Sec.~\ref{subsec:exp_sr_analysis}. The subsequent exposition mainly revolves around the construction of the shifting sequence $\{\eta_t\}_{t=1}^{T}$.

Equation~\eqref{eq:transit_0_t} indicates that the stochastic perturbation in state $\bm{x}_t$ is proportional to $\sqrt{\eta_t}$, incorporating a scaling factor $\kappa$. This observation motivates us to focus on the design of $\sqrt{\eta_t}$ rather than $\eta_t$. Previous work by Song \MakeLowercase{\textit{et al.}}~\cite{song2019generative} has suggested that 
maintaining a sufficiently small value for $\kappa\sqrt{\eta_1}$, such as 0.04 in LDM~\cite{rombach2022high}, is imperative to ensure $q(\bm{x}_1|\bm{x}_0,\bm{y}_0)\approx q(\bm{x}_0)$. Further considering $\eta_1 \to 0$, we set $\eta_1$ to be the minimum value between $(\sfrac{0.04}{\kappa})^2$ and $0.001$. For the terminal step $T$, we set $\eta_T$ as 0.999, guaranteeing $\eta_T \to 1$. For the intermediate timesteps $t \in [2, T-1]$, we propose a non-uniform geometric schedule for $\sqrt{\eta_t}$ as follows:
\begin{equation}
    \sqrt{\eta_t} = \sqrt{\eta_1} \times b_0^{\beta_t},~ t=2,\cdots,T-1,
    \label{eq:eta_schedule}
\end{equation}
where
\begin{subequations} \label{eq:hyper_schedule}
\begin{align}
   \beta_t &= \left(\frac{t-1}{T-1}\right)^p \times  (T-1), \label{eq:hyper_schedule_a} \\
   b_0 &=\exp\left[\frac{1}{2(T-1)}\log{\frac{\eta_T}{\eta_1}}\right].
   \label{eq:hyper_schedule_b}
   \end{align}
\end{subequations}
It should be noted that the choice of $\beta_t$ and $b_0$ is grounded in the assumption of $\beta_1=0$, $\beta_T=T-1$, and $\sqrt{\eta_T} = \sqrt{\eta_1} \times b_0^{T-1}$. The hyper-parameter $p$ controls the growth rate of $\sqrt{\eta_t}$, as depicted in Fig.~\ref{fig:schedule}(h).

The proposed noise schedule exhibits a high degree of flexibility in three key aspects. First, in the case of small values of $\kappa$, the final state $\bm{x}_T$ converges towards a perturbation around the LQ image, as illustrated in Fig.~\ref{fig:schedule}(c)-(d). Compared to the diffusion process ended at Gaussian noise, this design significantly shortens the length of the Markov chain, thereby improving the inference efficiency. Second, the hyper-parameter $p$ provides precise control over the shifting speed, enabling a fidelity-realism trade-off in the SR results as analyzed in Sec.~\ref{subsec:exp_sr_analysis}. Third, by setting $\kappa=40$ and $p=0.8$, our method achieves a diffusion process that degenerates into LDM~\cite{rombach2022high}. This is clearly demonstrated by the visual results of the diffusion process presented in Fig.~\ref{fig:schedule}(e)-(f), and further supported by the comparisons on the relative noise strength as shown in Fig.~\ref{fig:schedule}(g).
\begin{figure}
    \centering
    \includegraphics[width=\linewidth]{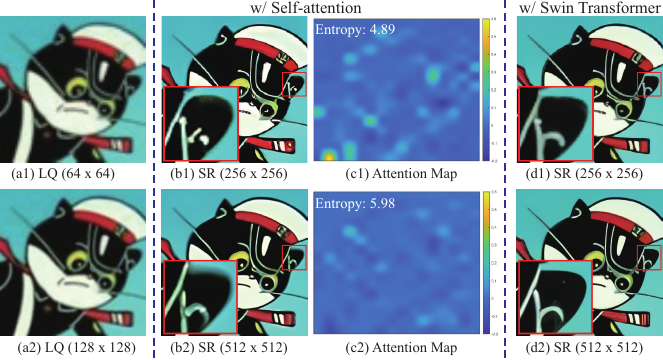}
    \caption{Visual comparison of two different models containing some self-attention layers (denoted as model-1) or Swin Transformers (denoted as model-2). (a1) and (a2): zoomed LQ images with resolutions of $64\times 64$ or $128\times 128$. (b1) and (b2): super-resolved results by model-1. (c1) and (c2): visualized attention maps extracted from the first self-attention layer of model-1. Note that these visualized results are obtained by first calculating the first principal component of PCA of the attention map and then reshaping it to the targeted size. In the left-upper corner, we annotate the entropy value of these attention maps. (d1) and (d2): super-resolved results by model-2. }
    \label{fig:atten-map}
\end{figure}

\subsection{Relation to Flow Matching} \label{subsec:flow-matching}
\zsyrevise{Flow matching~\cite{lipman2023flow}, also known as Rectified flow~\cite{liu2023flow}, is another advanced framework beyond diffusion models, focusing on finding the optimal transport map from one distribution to another. In this section, we present an alternative formulation of our proposed method through flow matching, offering a novel perspective on its theoretical foundation.}

\zsyrevise{
We first introduce two important definitions, namely the \textit{probability density path} $p_t:[0,1]\times \mathbb{R}^d \rightarrow \mathbb{R}_{>0}$, which is a time-dependent probability density function satisfying $\int p_t(x)\dif x=1$, and a time-dependent \textit{vector field} $v_t:[0,1]\times \mathbb{R}^d \rightarrow \mathbb{R}^d$. Given the data points $\bm{x}\in \mathbb{R}^d$, the vector field $v_t$ can be used to construct a time-dependent diffeomorphic map, called a \textit{flow} $\phi_t : [0,1]\times \mathbb{R}^d \rightarrow \mathbb{R}^d$, which is defined by the following ordinary differential equation:}
\begin{subequations}
\begin{align}
    \zsyrevise{\frac{\dif}{\dif t} \phi_t(\bm{x})} & \zsyrevise{= v_t(\phi_t(\bm{x})),} \\
    \zsyrevise{\phi_0(\bm{x})} & \zsyrevise{= \bm{x}.}
\end{align}
\end{subequations}

\zsyrevise{
Chen \textit{et al.}~\cite{chen2018neural} proposed to parameterize the vector field $v_t$ as a deep neural network $v_t(\cdot;\theta)$ with parameter $\theta$, leading to a deep parametric model of the flow $\phi_{t}$, called continuous \textit{normalizing flow} (NF).
In image generation, NF is often used to model the transport map between one simple known distribution $q_0$, typically Gaussian, and the data distribution $q_1$. Lipman~\textit{et al.}~\cite{lipman2023flow} developed the conditional flow matching technique that defines conditional probability path as follows:
\begin{equation}
    p_t(\bm{x}|\bm{x}_1) = \mathcal{N}(\bm{x}|\mu_t(\bm{x}_1), \sigma_t(\bm{x}_1)^2\bm{I}),
\end{equation}
where $\bm{x}_1 \sim q_1(\bm{x})$. Furthermore, the corresponding flow $\phi_t$ is specified in a simple format:
\begin{equation}
   \phi_t(\bm{x}) = \sigma_t(\bm{x}_1) \bm{x} + \mu_t(\bm{x}_1),  
\end{equation}
where $\bm{x} \sim q_0(x)$. This work provides a general framework with a close relationship to diffusion models and optimal transport theory, and more details can be found in~\cite{lipman2023flow,liu2022flow}.
}

\zsyrevise{
Even though our proposed method is formulated upon the diffusion model, it corresponds to a conditional flow between the LQ image distribution $q_0$ and the HQ image distribution $q_1$, specifically designed for IR. For a given image pair $\bm{y}_0$ and $\bm{x}_0$ from $q_0$ and $q_1$ respectively, the underlying probability path $p_t$ of our method can be expressed as:
\begin{equation}
     p_t(\bm{x}|\bm{x}_0) = \mathcal{N}(\bm{x}|\mu_t(\bm{x}_0), \sigma_t(\bm{x}_0)^2\bm{I}),
\end{equation}
where
\begin{equation}
     \mu_t(\bm{x}_0) = \eta_t \bm{x}_0 + (1-\eta_t)\bm{y}_0, ~
    \sigma_t(\bm{x}_0) = \kappa \sqrt{1-\eta_t},
\end{equation}
and $\eta_t$ is defined in Eq.~\eqref{eq:eta_schedule}. The conditional flow $\phi_t$ is a linear interpolation between the LQ and HQ images followed with a noise disturbance, i.e., 
\begin{equation}
    \phi(\bm{y}_0) =  \eta_t \bm{x}_0 + (1-\eta_t)\bm{y}_0 + \kappa \sqrt{1-\eta_t} \bm{\xi},
\end{equation}
where $\bm{y}_0 \sim q_0(\bm{y}_0)$, $\bm{\xi} \sim \mathcal{N}(\bm{0}, \bm{I})$. This intuitive and straightforward path provides a rapid transport map between LQ and HQ distributions, thereby improving the sampling inference significantly, aligning with the conclusions in~\cite{lipman2023flow,liu2023flow}.
}

\subsection{Discussion on Arbitrary Resolution}\label{subsec:discussion-swin}
It is widely acknowledged that the self-attention layer~\cite{vaswani2017attention}, a pivotal component in recent diffusion architectures, plays a crucial role in capturing global information in image generation. In the field of IR, however, it causes a blurring issue in handling the test images with arbitrary resolutions, particularly when the test resolution largely diverges from the training resolution. One typical example is provided in Figure~\ref{fig:atten-map}, considering two LQ images with different resolutions. The baseline model with multiple self-attention layers, which is trained on a resolution of $64\times 64$, performs well when the LQ image aligns with the training resolution but yields blurred results when confronted with a mismatched resolution of $128 \times 128$. 

To analyze the underlying reason, we visualize the attention maps extracted from the first attention layer in this baseline network, as shown in Fig.~\ref{fig:atten-map}(c1) and (c2). Note that these two attention maps are both interpolated to the resolution of $256\times 256$ for ease of comparison. Evidently, the example with a larger resolution tends to generate a more uniformly distributed attention map, i.e., Fig.~\ref{fig:atten-map}(c2), being consistent with the entropy\footnote{\zsyrevise{The average entropy of the attention map $\bm{W}\in\mathcal{R}^{n\times n}$ is defined as $-\frac{1}{n}\sum_{i=1}^{n}\sum_{j=1}^{n}w_{ij}\ln{w_{ij}}$, where we assume that each row of $\bm{W}$ represents the event probabilities of a discrete categorical distribution.}} values annotated on the left-upper corner\footnote{The principle of maximum entropy posits that it achieves the maximum entropy when the attention map conforms to a uniform distribution}. Consequently, a uniformly distributed attention map often leads to an over-smooth outcome, introducing undesirable distortions in performance.

To address this issue, some recent studies~\cite{whang2022deblurring,delbracio2023inversion} have chosen to discard the self-attention layers, a strategy that typically results in a noticeable decline in performance. Inspired by Liang et al.~\cite{liang2021swinir}, we propose a solution by substituting the self-attention layers with Swin Transformers~\cite{liu2021swin}. This straightforward replacement not only alleviates the blurring problem but also maintains the promised performance, as shown in Fig.~\ref{fig:atten-map}(d1) and (d2). This is because the Swin Transformer computes the attention map in a local window, thus being independent of the image resolution.

\section{Experiments on Image Super-resolution}\label{sec:exp-sr}
This section offers an evaluation of the proposed method on the task of image super-resolution (SR), with a particular focus on the setting of $\times4$ SR following existing studies~\cite{zhang2021designing,wang2021real}. We first provide ablation studies of the proposed model and then conduct a thorough comparison against recent state-of-the-art methods (SotAs) on one synthetic and two real-world datasets. For brevity in presentation, \textbf{\textit{our method is herein referred to ResShift or ResShiftL}}. The former is trained based on the primary loss in Eq.~\eqref{eq:loss_l2} with 15 diffusion steps, while the latter further introduces the perceptual regularization as shown in Eq.~\eqref{eq:loss_l2_lpips} with 4 steps. 

\subsection{Experimental Setup} \label{subsec:exp_sr_setup}
\vspace{2mm}\noindent\textbf{Training Details}. The HQ images in our training data, with a resolution of $256\times 256$, are randomly cropped from the training set of ImageNet~\cite{deng2009imagenet} like LDM~\cite{rombach2022high}. The LQ images are synthesized using the degradation pipeline of RealESRGAN~\cite{wang2021real}. To train our model, we adopted the Adam~\cite{kingma2015adam} algorithm with its default settings in PyTorch~\cite{paszke2019pytorch} and set the mini-batch size as 64. The learning rate is gradually decayed from $5\text{e-}5$ to $2\text{e-}5$ according to the annealing cosine schedule~\cite{loshchilov2017sgdr}, and a total of 500K iterations are implemented throughout the training. \zsyrevise{Our network is mainly built upon the UNet backbone in DDPM~\cite{ho2020denoising}, and the detailed architecture can be found in Fig.~\ref{fig:network} of the Appendix.} To increase our model's adaptability to arbitrary image resolutions, we replace the self-attention layer in Unet with Swin transformer~\cite{liu2021swin} as explained in Sec.~\ref{subsec:discussion-swin}.  

\vspace{2mm}\noindent\textbf{Test Datasets}. We randomly select 3000 images from the validation set of ImageNet~\cite{deng2009imagenet} as our synthetic test data, denoted as \textit{ImageNet-Test} for convenience. The LQ images are generated based on the commonly-used degradation model:
\begin{equation}
    \bm{y} = (\bm{x}*\bm{k})\downarrow + \bm{n}, \label{eq:degradation-test}
\end{equation}
where $\bm{k}$ is the blurring kernel, $\bm{n}$ is the noise,  $\bm{y}$ and $\bm{x}$ denote the LQ image and HQ image, respectively. To comprehensively evaluate the performance of our model, we consider more complicated types of blurring kernel, downsampling operator, and noise type. The detailed settings on them can be found in Appendix~\ref{subsec:degradation_app}. It should be noted that we select the HQ images from ImageNet~\cite{deng2009imagenet} instead of the prevailing datasets in SR, such as \textit{Set5}~\cite{bevilacqua2012low}, \textit{Set14}~\cite{zeyde2012single}, and \textit{Urban100}~\cite{huang2015single}. The rationale behind this setting is that these datasets only contain very few source images, which fail to thoroughly evaluate the performance of various methods under many different degradation types. 
\begin{table*}[t]
    \centering
    \caption{\zsyrevise{Quantitative comparisons of the proposed method with different attention layers on the synthetic dataset of \textit{ImageNet-Test} and the real-world dataset of \textit{RealSet80}.}}
    \label{tab:attention-abalation}
    \small
    \vspace{-2mm}
   \begin{tabular}{@{}C{2.2cm}@{}|@{}C{3.0cm}@{}| 
                   @{}C{1.6cm}@{} @{}C{1.6cm}@{} @{}C{1.7cm}@{} @{}C{2.0cm}@{} @{}C{2.0cm}@{}|
                   @{}C{2.0cm}@{} @{}C{2.0cm}@{}
                   }
         \Xhline{0.8pt}
         \multirow{2}*{Methods} & \multirow{2}*{Attention types} & \multicolumn{5}{c|}{\textit{ImageNet-Test}} & \multicolumn{2}{c}{\textit{RealSet80}}  \\
         \Xcline{3-9}{0.4pt}
         & & PSNR$\uparrow$ & SSIM$\uparrow$ & LPIPS$\downarrow$ & CLIPIQA$\uparrow$ & MUSIQ$\uparrow$ & CLIPIQA$\uparrow$ & MUSIQ$\uparrow$ \\
         \Xhline{0.4pt}
         Baseline  & Self-attention   & 24.97 & 0.6806 & 0.2137  & 0.5934 & 51.844 & 0.5883 & 59.090  \\
         ResShiftL & Swin Transformer & 25.02 & 0.6833 & 0.2076  & 0.5976 & 51.966 & 0.6418 & 61.022  \\
         \Xhline{0.8pt}
    \end{tabular}   
    \vspace{-2mm}
\end{table*}
\begin{table}[t]
    \centering
    \caption{\zsyrevise{Quantitative comparisons of our method with various fidelity losses ($L_1$ or $L_2$) on the \textit{ImageNet-Test} dataset.}}
    \label{tab:fidelity-loss}
    \small
    \vspace{-3mm}
   \begin{tabular}{@{}C{1.8cm}@{}|@{}C{1.3cm}@{} @{}C{1.2cm}@{} @{}C{1.3cm}@{} @{}C{1.7cm}@{} @{}C{1.5cm}@{}}
         \Xhline{0.8pt}
         \multirow{2}*{Methods}  & \multicolumn{5}{c}{Metrics} \\
         \Xcline{2-6}{0.4pt}
         & PSNR$\uparrow$ & SSIM$\uparrow$ & LPIPS$\downarrow$ & CLIPIQA$\uparrow$ & MUSIQ$\uparrow$ \\
         \Xhline{0.4pt}
         ResShiftL-L1  & 24.63 & 0.6710 & 0.2115  & 0.6261 & 53.8254  \\
         ResShiftL     & 25.02 & 0.6833 & 0.2076  & 0.5976 & 51.9656  \\
         \Xhline{0.4pt}
    \end{tabular}   
    \vspace{-2mm}
\end{table}
\begin{table}[t]
    \centering
    \caption{\zsyrevise{Quantitative comparisons of our method with ($\lambda=1$) or without ($\lambda=0$) perceptual loss on the \textit{ImageNet-Test} dataset.}}
    \label{tab:lpips-loss}
    \small
    \vspace{-3mm}
   \begin{tabular}{@{}C{2.4cm}@{}|@{}C{1.2cm}@{} @{}C{1.1cm}@{} @{}C{1.2cm}@{} @{}C{1.6cm}@{} @{}C{1.4cm}@{}}
         \Xhline{0.8pt}
         \multirow{2}*{Hyper-parameters}  & \multicolumn{5}{c}{Metrics} \\
         \Xcline{2-6}{0.4pt}
         & PSNR$\uparrow$ & SSIM$\uparrow$ & LPIPS$\downarrow$ & CLIPIQA$\uparrow$ & MUSIQ$\uparrow$ \\
         \Xhline{0.4pt}
         $\lambda=0$  & 25.64 & 0.6930 & 0.3242  & 0.4241 & 41.8308\\
         $\lambda=1$  & 25.02 & 0.6833 & 0.2076  & 0.5976 & 51.9656  \\
         \Xhline{0.4pt}
    \end{tabular}   
    \vspace{-2mm}
\end{table}

Two real-world datasets are adopted to evaluate the efficacy of our method. The first is \textit{RealSR}-V3~\cite{cai2019toward}, containing 100 real images captured by Canon 5D3 and Nikon D810 cameras. 
Additionally, we collect another real-world dataset named \textit{RealSet80}. It comprises 50 LQ images widely used in recent literature~\cite{martin2001database,matsui2017sketch,Ignatov2017DSLR,zhang2018ffdnet,wang2021real,lin2023diffbir}. The remaining 30 images are downloaded from the internet by ourselves.
\begin{figure}[t]
    \centering
    \includegraphics[width=1.0\linewidth]{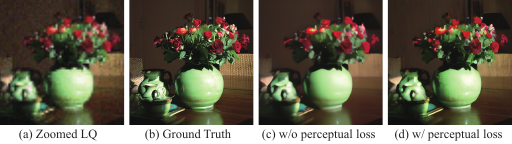}
    \vspace{-6mm}
    \caption{\zsyrevise{Ablation studies of our method regarding the perceptual loss. 
    }}
    \label{fig:ablation-lpips}
\end{figure}
\begin{table}[t]
    \centering
    \caption{\zsyrevise{Quantitative results and the corresponding standard deviation (std) of the proposed method under multiple random seeds on the dataset of \textit{ImageNet-Test}.}}
    \label{tab:randomness}
    \small
    \vspace{-3mm}
   \begin{tabular}{@{}C{1.6cm}@{}|@{}C{1.4cm}@{} @{}C{1.4cm}@{} @{}C{1.4cm}@{}
                    @{}C{1.4cm}@{}|@{}C{1.4cm}@{}}
         \Xhline{0.8pt}
         Metrics         & Seed-1 & Seed-2 & Seed-3 & Seed-4 & Std  \\
         \Xhline{0.4pt}
         PSNR $\uparrow$ & 25.02 & 25.01 & 25.03 & 25.01  &0.00829 \\
         SSIM $\uparrow$ & 0.6833 & 0.6826 & 0.6834 & 0.6830  &0.00031 \\
         LPIPS $\downarrow$ & 0.2076 & 0.2074 & 0.2076 & 0.2075  &0.00008 \\
         \Xhline{0.4pt}
    \end{tabular}   
    \vspace{-2mm}
\end{table}
\begin{table*}[t]
    \centering
    \caption{\zsyrevise{Quantitative comparison on performance, running time, and the number of parameters of different methods on \textit{ImageNet-Test} dataset for Image Super-resolution. The results of the diffusion-based methods are denoted as ``Method-A'', where ``A'' represents the number of sampling steps. Running time is tested on NVIDIA Tesla A100 GPU on the x4 (64$\rightarrow$256) SR task. 
    The non-trainable parameters, such as the parameters of VQGAN in LDM, are marked with \textcolor[gray]{0.5}{gray} color for clarity.}}
    \label{tab:imagenet_testing_sr}
    \small
    \vspace{-2mm}
    \begin{tabular}{@{}C{3.0cm}@{}|
                    @{}C{1.6cm}@{} @{}C{1.6cm}@{}@{}C{2.0cm}@{}  @{}C{2.0cm}@{} @{}C{2.0cm}@{}  @{}C{2.2cm}@{} @{}C{2.8cm}@{}}
        \Xhline{0.8pt}
        \multirow{2}*{Methods} & \multicolumn{5}{c}{Metrics} \\
        \Xcline{2-8}{0.4pt}
            & PSNR$\uparrow$ & SSIM$\uparrow$ & LPIPS$\downarrow$ & CLIPIQA$\uparrow$ & MUSIQ$\uparrow$ & Runtime (s) & \#Params (M) \\
            \Xhline{0.4pt}
            ESRGAN~\cite{wang2018esrgan}     & 20.67 & 0.448 & 0.485 & 0.451 & 43.615 & 0.038  & 16.70  \\
            RealSR-JPEG~\cite{ji2020real}    & 23.11 & 0.591 & 0.326 & 0.537 & 46.981 & 0.038  & 16.70  \\
            BSRGAN~\cite{zhang2021designing} & 24.42 & 0.659 & 0.259 & 0.581 & 54.697 & 0.038  & 16.70  \\
            SwinIR~\cite{liang2021swinir}    & 23.99 & 0.667 & 0.238 & 0.564 &53.790  & 0.107  & 28.01 \\
            RealESRGAN~\cite{wang2021real}   & 24.04 & 0.665 & 0.254 & 0.523 & 52.538 & 0.038  & 16.70 \\
            DASR~\cite{liang2022efficient}   & 24.75 & 0.675 & 0.250 & 0.536 & 48.337 & 0.022  & 8.06  \\
            DiffIR-4~\cite{Xia_2023_ICCV}    & 24.50 & 0.674 &0.217  & 0.554 & 54.567 & 0.161 & 26.48 \\
            \Xhline{0.4pt}
            LDM-50~\cite{rombach2022high}    & 24.17 & 0.637 & 0.245 & 0.600 & 52.665 & 0.773  & \multirow{3}*{113.60\textcolor[gray]{0.5}{+55.32}}   \\
            LDM-15~\cite{rombach2022high}    & 24.89 & 0.670 & 0.269 & 0.512 & 46.419  & 0.247  &  \\
            LDM-4~\cite{rombach2022high}     & 24.74 & 0.657 & 0.345 & 0.372 & 38.161  & 0.077 & \\
            \Xhline{0.4pt}
            StableSR-50~\cite{wang2023exploiting}    & 22.96 & 0.611 & 0.264 & 0.666 & 59.559  & 3.205 & \multirow{3}*{152.70\textcolor[gray]{0.5}{+1422.49}}\\
            StableSR-15~\cite{wang2023exploiting}    & 23.37 & 0.631 & 0.262 & 0.660 & 59.492 & 1.070 &  \\
            StableSR-4~\cite{wang2023exploiting}     & 24.11 & 0.658 & 0.287 & 0.580 & 53.698 & 0.399 &  \\
            \Xhline{0.4pt}
            CCSR-45~\cite{sun2023improving}    & 24.67 & 0.661 & 0.236       & 0.614 & 58.242 & 4.500 & \multirow{3}*{363.15\textcolor[gray]{0.5}{+1303.60}}\\
            CCSR-15~\cite{sun2023improving}    & 24.86 & 0.669 & 0.243 & 0.581 & 55.773  & 1.670 & \\
            CCSR-4~\cite{sun2023improving}    & 25.37 & 0.694 & 0.282  & 0.450 & 46.204  & 0.622 & \\
            \Xhline{0.4pt}
            ResShift-15 & 25.01 & 0.677   & 0.231 & 0.592 & 53.660 & 0.682 &\multirow{2}*{118.59\textcolor[gray]{0.5}{+55.32}} \\
            ResShiftL-4             & 25.02      & 0.683 & 0.208 & 0.598 & 51.966 &0.186 & \\   
       \Xhline{0.8pt}
    \end{tabular} 
    \vspace{-2mm}
\end{table*}
\begin{figure*}[t]
    \centering
    \includegraphics[width=\linewidth]{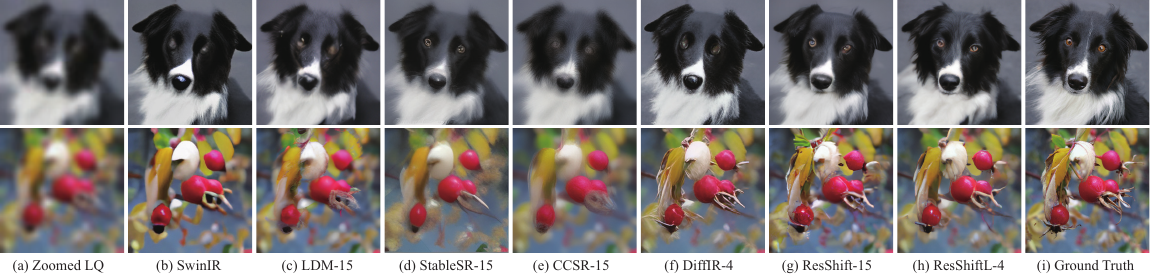}
    \vspace{-6mm}
    \caption{\zsyrevise{Qualitative results of different methods on the synthetic \textit{ImageNet-Test} dataset for image super-resolution. Note that we only display the comparison results to the recent five SotA methods in (b)-(f) due to the page limitation, and the complete results are presented in Fig.~\ref{fig:syn_imagenet_app} of the Appendix.}}
    \label{fig:syn_imagenet}
    \vspace{-3mm}
\end{figure*}
\begin{table}[t]
    \centering
    \caption{\zsyrevise{Quantitative results of different methods on two real-world datasets for image super-resolution. Note that the results of diffusion-based methods are denoted as ``Method-A'', where ``A'' represents the number of sampling steps. 
    }}
    \label{tab:real_testing}
    \small
    \vspace{-2mm}
    \begin{tabular}{@{}C{2.6cm}@{}|
                    @{}C{1.6cm}@{} @{}C{1.5cm}@{}| 
                    @{}C{1.6cm}@{} @{}C{1.5cm}@{} }
        \Xhline{0.8pt}
        \multirow{3}*{Methods} & \multicolumn{4}{c}{Datasets} \\
        \Xcline{2-5}{0.4pt}
            & \multicolumn{2}{c|}{\textit{RealSR}-V3~\cite{cai2019toward}}  & \multicolumn{2}{c}{\textit{RealSet80}} \\
            \Xcline{2-5}{0.4pt}
            & CLIPIQA$\uparrow$ & MUSIQ$\uparrow$   
            & CLIPIQA$\uparrow$ & MUSIQ$\uparrow$  \\
            \Xhline{0.4pt}
            ESRGAN~\cite{wang2018esrgan}     & 0.2362 & 29.048  & 0.4165 & 48.153   \\
            RealSR-JPEG~\cite{ji2020real}    & 0.3615 & 36.076  & 0.5828 & 57.379  \\
            BSRGAN~\cite{zhang2021designing} & 0.5439 & 63.586  & 0.6263 & 66.629 \\
            SwinIR~\cite{liang2021swinir}    & 0.4654 & 59.632  & 0.6072 & 64.739 \\
            RealESRGAN~\cite{wang2021real}   & 0.4898 & 59.678  & 0.6189 & 64.496  \\
            DASR~\cite{liang2022efficient}   & 0.3629 & 45.825  & 0.5311 & 58.974   \\
            DiffIR-4~\cite{Xia_2023_ICCV}    & 0.4315 & 57.449  & 0.5909 & 62.028 \\
            \Xhline{0.4pt}
            LDM-50~\cite{rombach2022high}    & 0.4907 & 54.391 & 0.5511 & 55.826  \\
            LDM-15~\cite{rombach2022high}    & 0.3836 & 49.317 & 0.4592  &50.972 \\
            LDM-4~\cite{rombach2022high}     & 0.2865 & 43.205 & 0.3582  &45.182 \\
             \Xhline{0.4pt}
            StableSR-50~\cite{wang2023exploiting}    & 0.5208 & 60.177 & 0.6214 & 62.761  \\
            StableSR-15~\cite{wang2023exploiting}    & 0.4974 & 59.099  & 0.5975      & 61.476  \\
            StableSR-4~\cite{wang2023exploiting}     & 0.4392 & 56.179  & 0.5250      & 57.445  \\
            \Xhline{0.4pt}
            CCSR-45~\cite{sun2023improving}  & 0.5681 & 63.222 & 0.6385 & 65.889 \\
            CCSR-15~\cite{sun2023improving}  & 0.5540 & 62.331 & 0.6284 & 64.859 \\
            CCSR-4~\cite{sun2023improving}   & 0.4893 & 58.039 & 0.5550 & 59.646 \\
             \Xhline{0.4pt}
            ResShift-15              & 0.5958 & 58.475 & 0.6645 & 62.782  \\
            ResShiftL-4              & 0.5995 & 57.554 & 0.6418 & 61.022    \\
       \Xhline{0.4pt}
    \end{tabular} 
\end{table}

\vspace{2mm}\noindent\textbf{Compared Methods}. We evaluate the effectiveness of ResShift and ResShiftL in comparison to nine recent SR methods, namely RealSR-JPEG~\cite{ji2020real}, BSRGAN~\cite{zhang2021designing}, RealESRGAN~\cite{wang2021real}, SwinIR~\cite{liang2021swinir}, DASR~\cite{liang2022efficient}, LDM~\cite{rombach2022high}, DiffIR~\cite{Xia_2023_ICCV}, StableSR~\cite{wang2023exploiting}, and CCSR~\cite{sun2023improving}. 
For a fair comparison, we accelerate the diffusion-based methods, including LDM, DiffIR, StableSR, and CCSR, to 15 or 4 steps using their default accelerating algorithm during inference. 
For clarity, the results of these diffusion-based methods are denoted as ``Method-A", where ``A" represents the number of inference steps. 

\vspace{2mm}\noindent\textbf{Evaluation Metrics}. We evaluate the efficacy of different methods using five widely used metrics, including PSNR, SSIM~\cite{zhou2004image}, LPIPS~\cite{zhang2018unreasonable}, CLIPIQA~\cite{wang2022exploring}, and MUSIQ~\cite{ke2021musiq}. 
Notably, CLIPIQA and MUSIQ are both non-reference metrics specifically designed for assessing the realism of images. Particularly, CLIPIQA leverages the CLIP~\cite{radford2021learning} model, pre-trained on the extensive Laion400M~\cite{schuhmann2021laion} dataset, thereby demonstrating strong generalization ability. 
\begin{figure*}[t]
    \centering
    \includegraphics[width=\linewidth]{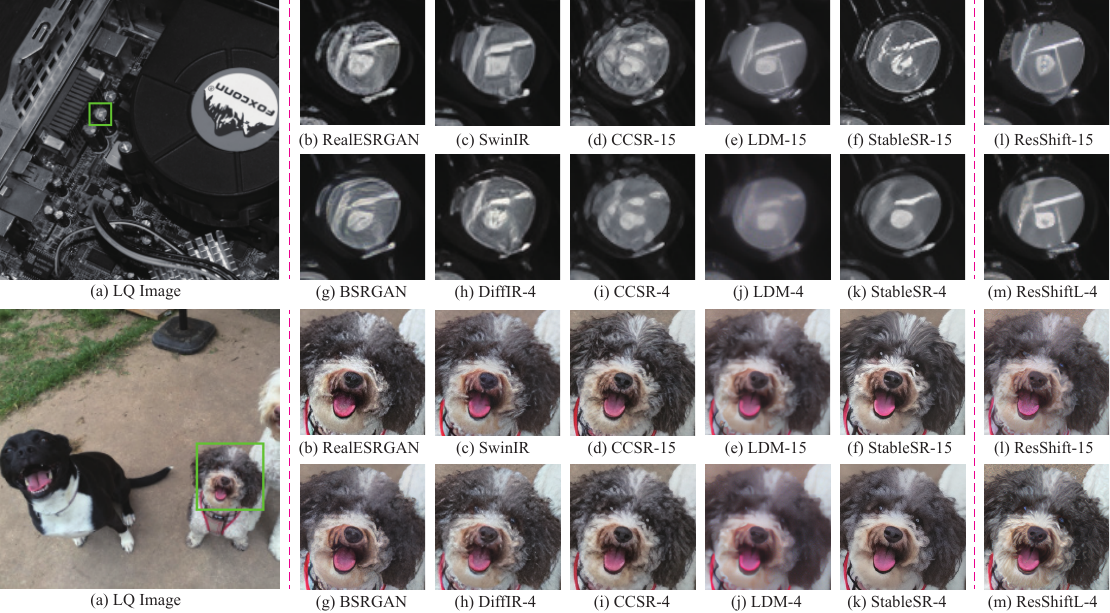}
    \vspace{-6mm}
    \caption{Qualitative comparisons on three real-world examples from \textit{RealSet80}. Please zoom in for a better view.}
    \label{fig:real_data}
\end{figure*}

\subsection{Ablation Studies} \label{subsec:exp_sr_analysis}
\zsyrevise{In this part, we provide some necessary ablation studies on several components in our model. More comprehensive analysis about the noise schedule, perception-distortion trade-off, and comparisons with more advanced samplers can be found in Appendix~\ref{subsubsec:analysis_sr_app}.}

\vspace{2mm}\noindent\textbf{Fidelity loss}. \zsyrevise{The loss function of our method incorporates both a fidelity loss and a perceptual regularizer, as shown in Eq.~\eqref{eq:loss_l2_lpips}. The fidelity loss is formulated as the $L_2$ norm, quantifying the discrepancy between the predicted HQ image and the underlying ground truth. We have also explored the use of an $L_1$ norm in place of the $L_2$ norm for the fidelity loss, resulting in a variant of our model denoted as \textit{ResShiftL-L1}. Comparison results are summarized in Table~\ref{tab:fidelity-loss}, demonstrating that ResShiftL outperforms ResShiftL-L1 on reference metrics, while ResShiftL-L1 shows superior performance on non-reference metrics. Considering the high fidelity requirement for IR, we adopted the $L_2$ norm in this study.}

\vspace{2mm}\noindent\textbf{Perceptual loss}. \zsyrevise{In contrast to our conference version~\cite{yue2023resshift}, this study integrates an additional perceptual regularizer, detailed in Eq.~\eqref{eq:loss_l2_lpips}, which enhances the model efficiency by reducing the sampling steps from 15 to 4. The ablation study summarized in Table~\ref{tab:lpips-loss} indicates that while the introduction of the perceptual loss results in a slight decrease in PSNR and SSIM, it yields significant improvements in LPIPS, CLIPIQA, and MUSIQ. These latter three metrics more truthfully reflect the perceptual quality and realism of images, as supported by the visual comparisons in Fig.~\ref{fig:ablation-lpips}. Therefore, considering both performance and efficiency, the incorporation of the perceptual regularizer proves to be a critical enhancement.}

\vspace{2mm}\noindent\textbf{Swin Transformer}. \zsyrevise{As discussed in Sec.~\ref{subsec:discussion-swin}, we replace the self-attention layers in the diffusion Unet with Swin Transformer blocks to address the arbitrary resolution issue. 
Table~\ref{tab:attention-abalation} provides a quantitative comparison of this modification. On the synthetic \textit{ImageNet-Test} dataset, where both training and testing images are of consistent resolution, models with either self-attention layers or Swin Transformer blocks exhibit comparable performance. In contrast, on the real-world dataset \textit{RealSet80}, which contains images of varying resolutions, the baseline model using self-attention layers suffers from a significant performance drop. This is mainly attributed to the inability of self-attention layers to generalize across resolutions that largely deviate from those encountered during training. A more comprehensive analysis and visualization from the perspective of information entropy are presented in Sec.~\ref{subsec:discussion-swin} and Fig.~\ref{fig:atten-map}.}

\vspace{2mm}\noindent\textbf{Sampling Randomness}. \zsyrevise{We discuss the sampling randomness caused by the stochastic sampling of diffusion models within the task of IR. Firstly, IR is an ill-posed problem, particularly in severely degraded scenarios where multiple HQ outputs can correspond to a single LQ image. The random sampling mechanism of diffusion models facilitates a one-to-many mapping, effectively addressing this ill-posed issue by generating diverse but plausible restoration outcomes for any testing image. Secondly, high fidelity is crucial for the task of IR. Our proposed method designs a diffusion process between the HQ and LQ images, rather than starting from random Gaussian noise, which reduces the randomness of sampling to a certain extent. Additionally, Table~\ref{tab:randomness} lists the quantitative comparisons of our method under various random seeds, and corresponding visual results can be found in Fig.~\ref{fig:randomness-sr} of the Appendix. These results empirically demonstrate the consistency across different outputs. On the other hand, while some level of randomness is present, it is manageable and beneficial for handling the ill-posedness of IR.}

\begin{table*}[t]
    \centering
    \caption{Quantitative comparisons of various methods on the test dataset \textit{ImageNet-Test} for inpainting. The best and second best results are highlighted in \textbf{bold} and \underline{underline}, respectively.}
    \label{tab:metric_inpainting_imagenet}
    \vspace{-3mm}
    \small
    \begin{tabular}{@{}C{1.8cm}@{}| @{}C{1.7cm}@{}|
                    @{}C{2.2cm}@{} @{}C{2.0cm}@{} @{}C{2.0cm}@{} @{}C{2.1cm}@{}
                    @{}C{2.4cm}@{} @{}C{1.9cm}@{} @{}C{2.0cm}@{} }
        \Xhline{0.8pt}
         \multirow{2}*{Mask Types}  &\multirow{2}*{Metrics} & \multicolumn{7}{c}{Methods} \\
         \Xcline{3-9}{0.4pt}
              & & DeepFillv2~\cite{yu2019free}   & LaMa~\cite{suvorov2022resolution}   & RePaint~\cite{Lugmayr2022repaint}   & DDRM~\cite{kawar2022denoising}   & Score-SDE~\cite{song2021scorebased}   & MCG~\cite{chung2022improving} &  ResShiftL \\
        \Xhline{0.4pt}
        \multirow{2}*{Box} & LPIPS$\downarrow$  & 0.1524  & \underline{0.1158}  & 0.1498   & 0.2241   & 0.2073  & 0.1464  & \textbf{0.1156}  \\
            & CLIPIQA$\uparrow$  & 0.4539  & 0.4492  & 0.4586   & \textbf{0.4705}    & 0.4350  & \underline{0.4639}   & 0.4587 \\
        \hline 
        \multirow{2}*{Irregular} & LPIPS$\downarrow$  & 0.2523  & \underline{0.1959}  & 0.2569   & 0.3712   & 0.3350  & 0.2389  & \textbf{0.1931}  \\
            & CLIPIQA$\uparrow$  & 0.4199  & 0.4204  & \underline{0.4392}   & 0.4304    & 0.4131  & 0.4388   & \textbf{0.4432}  \\
            \hline 
        \multirow{2}*{Half} & LPIPS$\downarrow$  & 0.3237  & \underline{0.2925}  & 0.3331   & 0.4404   & 0.3709  & 0.3120  & \textbf{0.2663}   \\
            & CLIPIQA$\uparrow$  & 0.4147  & 0.4183  & \underline{0.4490}   & 0.4316    & 0.4263  & \textbf{0.4599}   & 0.4476 \\
             \hline 
        \multirow{2}*{Expand} & LPIPS$\downarrow$  & 0.5032  & \underline{0.3561}  & 0.4957   & 0.6081   & 0.5620  & 0.4320  & \textbf{0.3439}  \\
            & CLIPIQA$\uparrow$  & 0.4480  & 0.4251  & 0.4530   & 0.4276    & 0.4293  & \textbf{0.4611}   & \underline{0.4581}  \\
            \hline
        \multirow{2}*{Average} & LPIPS$\downarrow$  & 0.2914  & \underline{0.2401}  & 0.3089   & 0.4152   & 0.3688  & 0.2823  & \textbf{0.2298}  \\
            & CLIPIQA$\uparrow$  & 0.4310  & 0.4282  & 0.4499   & 0.4400    & 0.4260  & \textbf{0.4559}   & \underline{0.4519} \\
        \Xhline{0.8pt}
    \end{tabular}
\end{table*}
\begin{table*}[t]
    \centering
    \caption{Quantitative comparisons of various methods on the test dataset \textit{CelebA-Test} for inpainting. The best and second best results are highlighted in \textbf{bold} and \underline{underline}, respectively.}
    \label{tab:metric_inpainting_celeba}
    \vspace{-3mm}
    \small
    \begin{tabular}{@{}C{1.8cm}@{}| @{}C{1.7cm}@{}|
                    @{}C{2.2cm}@{} @{}C{2.0cm}@{} @{}C{2.0cm}@{} @{}C{2.1cm}@{}
                    @{}C{2.4cm}@{} @{}C{1.9cm}@{} @{}C{2.0cm}@{} }
        \Xhline{0.8pt}
         \multirow{2}*{Mask Types}  &\multirow{2}*{Metrics} & \multicolumn{7}{c}{Methods} \\
         \Xcline{3-9}{0.4pt}
              & & DeepFillv2~\cite{yu2019free}   & LaMa~\cite{suvorov2022resolution}   & RePaint~\cite{Lugmayr2022repaint}   & DDRM~\cite{kawar2022denoising}   & Score-SDE~\cite{song2021scorebased}   & MCG~\cite{chung2022improving} &  ResShiftL \\
        \Xhline{0.4pt}
        \multirow{2}*{Box} & LPIPS$\downarrow$  & 0.0719  & \textbf{0.0533}  & 0.0702   & 0.0755   & 0.1087  & 0.0764  & \underline{0.0550}  \\
            & CLIPIQA$\uparrow$  & 0.4487  & 0.4365  & \underline{0.4754}   & 0.4521    & 0.4547  & 0.4714   & \textbf{0.4915} \\
        \hline 
        \multirow{2}*{Irregular} & LPIPS$\downarrow$  & 0.1690  & \underline{0.1221}  & 0.1602   & 0.1632   & 0.2315  & 0.1522  & \textbf{0.1169}  \\
            & CLIPIQA$\uparrow$  & 0.4297  & 0.4214  & 0.4558   & 0.4359    & 0.4385  & \underline{0.4649}   & \textbf{0.5029}  \\
            \hline 
        \multirow{2}*{Half} & LPIPS$\downarrow$  & 0.2147  & \underline{0.1603}  & 0.1936   & 0.2039   & 0.2415  & 0.1853  & \textbf{0.1535}   \\
            & CLIPIQA$\uparrow$  & 0.4129  & 0.4056  & 0.4751   & 0.4424    & 0.4603  & \underline{0.4772}   & \textbf{0.5189} \\
             \hline 
        \multirow{2}*{Expand} & LPIPS$\downarrow$  & 0.4003  & \underline{0.2961}  & 0.3858   & 0.3978   & 0.4456  & 0.3471  & \textbf{0.2772}  \\
            & CLIPIQA$\uparrow$  & 0.3989  & 0.4053  & 0.4469   & 0.4280    & 0.4378  & \underline{0.5022}   & \textbf{0.5111}  \\
            \hline
        \multirow{2}*{Average} & LPIPS$\downarrow$  & 0.2140  & \underline{0.1580}  & 0.2029   & 0.2101   & 0.2568  & 0.1902  & \textbf{0.1506}  \\
            & CLIPIQA$\uparrow$  & 0.4225  & 0.4172  & 0.4633   & 0.4396    & 0.4478  & \underline{0.4789}   & \textbf{0.5061} \\
        \Xhline{0.8pt}
    \end{tabular}
\end{table*}

\subsection{Evaluation on Synthetic Data}
We present a comparative analysis of the proposed method with recent SotA approaches on the \textit{ImageNet-Test} dataset, as summarized in Table~\ref{tab:imagenet_testing_sr} and Fig.~\ref{fig:syn_imagenet}. This evaluation reveals the following conclusions: i) Diffusion-based methods demonstrate significant advantages in terms of non-reference metrics; however, their performance on reference metrics is hindered by the inherent randomness in the sampling procedure. ii) Among diffusion-based methods, our proposed method exhibits superior performance across both reference and non-reference metrics with the same number of sampling steps, indicating an improved fidelity-realism trade-off. iii) ResShiftL is notably faster than other diffusion-based methods, achieving a preeminent balance between performance and efficiency. Even in comparison with the SotA GAN-based method SwinIR~\cite{liang2021swinir}, it not only maintains comparable speed but also delivers superior performance. This efficiency is attributed to our well-designed diffusion model, which has a shorter transition trajectory. 

\subsection{Evaluation on Real-World Data}
Table~\ref{tab:real_testing} lists the comparative evaluation using CLIPIQA~\cite{wang2022exploring} and MUSIQ~\cite{ke2021musiq} for various approaches on two real-world datasets, namely \textit{RealSR}-V3~\cite{cai2019toward} and \textit{RealSet80}. Note that CLIPIQA, benefiting from the powerful representative capability inherited from CLIP, performs consistently and robustly in assessing the perceptional quality of natural images. The results in Table~\ref{tab:real_testing} reveal that the proposed ResShift or ResShiftL notably outperforms existing methods in terms of CLIPIQA. This suggests that the restored outputs by our method better align with human visual and perceptive systems. In the case of MUSIQ evaluation, ResShift attains competitive performance when compared to current SotA methods, namely BSRGAN~\cite{zhang2021designing}, SwinIR~\cite{liang2021swinir}, and RealESRGAN~\cite{wang2021real}. Collectively, our method shows promising capability in addressing real-world SR challenges.

We display three real-world examples in Fig.~\ref{fig:real_data}. We consider diverse scenarios, including text, animal, and natural images to ensure a comprehensive evaluation. An obvious observation is that ResShift or ResShiftL produces more naturalistic image structures. We note that the recovered results of LDM~\cite{rombach2022high} and StableSR~\cite{wang2023exploiting} are excessively smooth when compressing the inference steps to match with our proposed method, i.e., 15 or 4 steps, largely deviating from the training procedure's 1,000 steps. 
Even though other GAN-based methods may also succeed in hallucinating plausible structures to some extent, they are often accompanied by obvious artifacts.
\begin{figure*}[t]
    \centering
    \includegraphics[width=\linewidth]{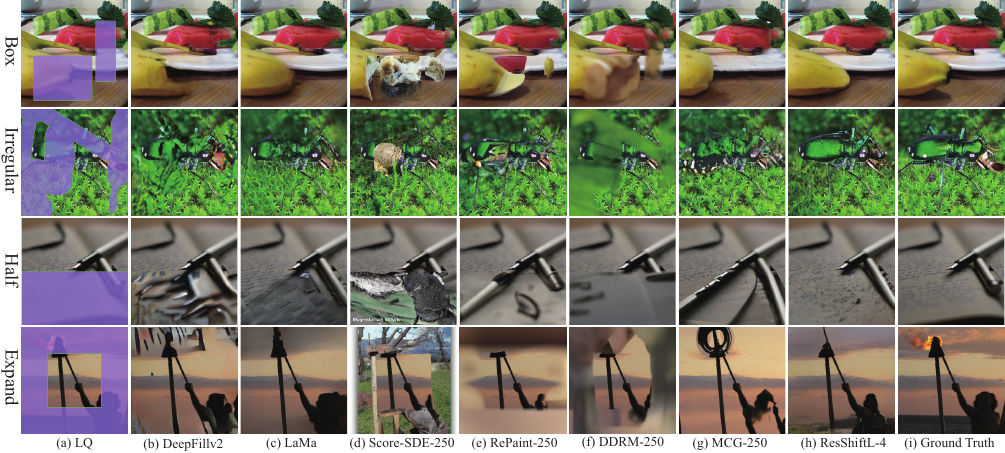}
    \vspace{-6mm}
    \caption{Visual comparisons of various methods on the test dataset of \textit{ImageNet-Test} for inpainting. The results of diffusion-based methods are denoted as ``Method-A'', where ``A'' represents the number of sampling steps. The masked areas are highlighted using a purple color. Please zoom in for a better view.}
    \label{fig:imagenet_inpainting}
\end{figure*}

\section{Experiments on Image Inpainting}\label{sec:exp-inpainting}
The proposed diffusion model is a general framework for IR. This section presents a series of experiments to validate its effectiveness in the task of image inpainting. Additional experimental results on blind face restoration and image deblurring are provided in Appendix~\ref{subsec:exp_bfr_app}.

\subsection{Experimental Setup}
\noindent\textbf{Training Details}. In addressing the task of inpainting, we train two variants of the ResShiftL model, both implemented at a resolution of $256\times 256$. These two variants are tailored for natural images and facial images, respectively. The former model is trained using the training dataset of ImageNet~\cite{deng2009imagenet}, while the latter is trained on the widely used face dataset FFHQ~\cite{karras2019style}. During training, we randomly generate the image masks to synthesize the LQ images following LaMa~\cite{suvorov2022resolution}. The other training configurations are kept consistent with those in image super-resolution, as detailed in Sec.~\ref{subsec:exp_sr_setup}.
\begin{figure*}[t]
    \centering
    \includegraphics[width=\linewidth]{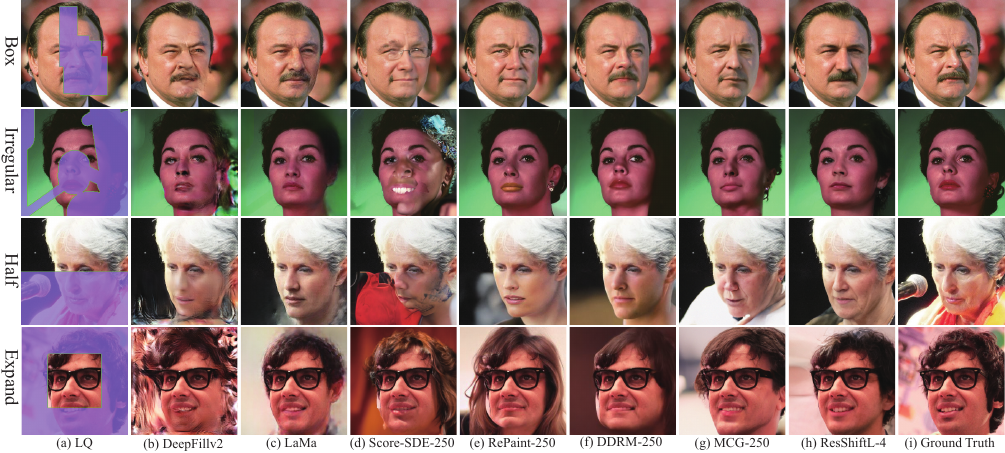}
    \vspace{-6mm}
    \caption{Visual comparisons of various methods on the test dataset of \textit{CelebA-Test} for inpainting. The results of diffusion-based methods are denoted as ``Method-A'', where ``A'' represents the number of sampling steps. The masked areas are highlighted using a purple color. Please zoom in for a better view.}
    \label{fig:celeba_inpainting}
\end{figure*}
\begin{figure*}[t]
    \centering
    \includegraphics[width=\linewidth]{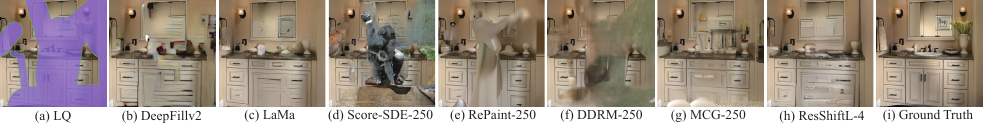}
    \vspace{-6mm}
    \caption{\zsyrevise{One typically failed example on the task of natural image inpainting.}}
    \label{fig:failed_inpainting}
\end{figure*}

\vspace{2mm}\noindent\textbf{Test Datasets}. Two test datasets are constructed by randomly selecting 2,000 images from the validation dataset of ImageNet~\cite{deng2009imagenet} and CelebA-HQ~\cite{karras2018progressive}, to facilitate an assessment on natural images and facial images, respectively. These images in each dataset are uniformly divided into four groups to synthesize different types of masked images. To ensure a thorough evaluation, four distinct mask types, denoted as ``Box'' mask, ``Irregular'' mask, ``Half'' mask, and ``Expand'' mask, are considered as visually shown in Fig.~\ref{fig:imagenet_inpainting}. For each mask type, we randomly generate a set of 500 masks, and then employ them to simulate the LQ images. These two datasets are denoted as \textit{ImageNet-Test} and \textit{CelebA-Test} in this section.

\vspace{2mm}\noindent\textbf{Compared Methods}. In order to evaluate the efficacy of ResShiftL, a comparative analysis is conducted against two GAN-based methods, including DeepFillv2~\cite{yu2019free} and LaMa~\cite{suvorov2022resolution}, as well as four diffusion-based methods, namely Score-SDE~\cite{song2021scorebased}, RePaint~\cite{Lugmayr2022repaint}, DDRM~\cite{kawar2022denoising}, and MCG~\cite{chung2022improving}. For the diffusion-based methods, we accelerate their sampling process to 250 steps using the DDIM~\cite{song2021denoising} algorithm.

\vspace{2mm}\noindent\textbf{Evaluation Metrics}. For the sake of comprehensively assessing the performance of various approaches, we adopt one full-reference metric LPIPS~\cite{zhang2018unreasonable} and one no-reference metric CLIPIQA~\cite{wang2022exploring} as our principal evaluative criteria.

\subsection{Comparison with SotA Methods}
We provide a quantitative evaluation of different methods on the test dataset of \textit{ImageNet-Test} and \textit{CelebA-Test}, as detailed in Table~\ref{tab:metric_inpainting_imagenet} and Table~\ref{tab:metric_inpainting_celeba}, respectively. The proposed ResShiftL achieves the best or, at the very least, comparable performance to recent SotA methods across most cases, particularly excelling in the more challenging mask types such as ``Irregular'', ``Half'', and ``Expand''. In comparison to other diffusion-based approaches, ResShiftL still maintains a competitive advantage, even with a significantly reduced number of sampling steps (4 vs. 250). 

A series of visual illustrations on various mask types are displayed in Fig.~\ref{fig:imagenet_inpainting} and Fig.~\ref{fig:celeba_inpainting}. In the case of ``Box'' mask, recent methods, namely LaMa~\cite{suvorov2022resolution} and MCG~\cite{chung2022improving}, and our ResShiftL all perform well. For the other three mask types containing large occluded areas, most of the comparison methods fail to handle such complicated scenarios. In contrast, the proposed ResShiftL consistently yields more plausible and realistic results under these scenarios, especially on the preservation of coherency to the unmasked regions. The qualitative analysis presented herein reaffirms the stability and exceptional performance of ResShiftL, aligning with the quantitative comparison above. 

\zsyrevise{While ResShiftL has demonstrated strong performance in most scenarios, failed examples still exist, particularly in cases involving large masked areas, as illustrated in Fig.~\ref{fig:failed_inpainting}. Existing methods struggle to effectively deal with such an extremely occluded example, mainly because the available information in this image is too limited. To address this challenge, a potential improvement avenue is introducing more supplementary guidance, such as text prompts. We leave the exploration in this direction for future research.}
\section{Conclusion} \label{sec:conclusion}
In this work, we have introduced an efficient diffusion model specifically designed for IR. Unlike existing diffusion-based IR methods that require a large number of iterations to achieve satisfactory results, our proposed method is capable of formulating a diffusion model with only four sampling steps, thereby significantly improving the efficiency during inference. The core idea is to corrupt the HQ image towards its LQ counterpart instead of the Gaussian white noise. This strategy can effectively truncate the length of the diffusion model. Extensive experiments on the tasks of image super-resolution and image inpainting have demonstrated the superiority of our proposed method. In addition, more discussion on the limitations of the proposed method can be found in the Appendix. We believe that our work will pave the way for the development of more efficient and effective diffusion models to address the IR problem.

 
\bibliography{reference}
\bibliographystyle{IEEEtran}

\begin{IEEEbiography}[{\includegraphics[width=1in,height=1.25in,clip,keepaspectratio]{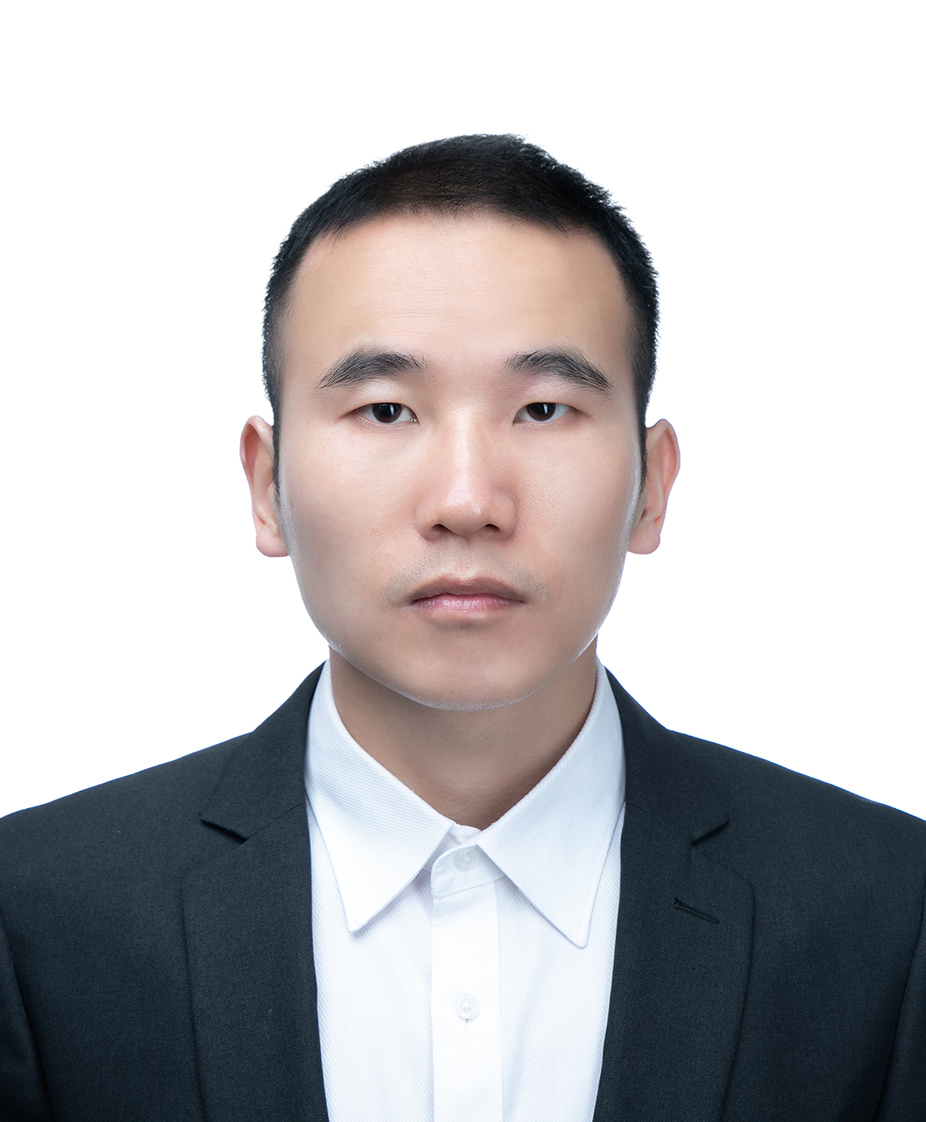}}]{Zongsheng Yue}
(Member, IEEE) received his Ph.D. degree from Xi'an Jiaotong University, Xi'an, China, in 2021. He is currently a postdoctoral research fellow with the College of Computing and Data Science at Nanyang Technological University. From September 2021 to March 2022, he was an associate researcher in the Department of Computer Science at Hong Kong University. He was a research assistant at the Department of Computing, Hong Kong Polytechnic University, from October 2018 to June 2019, and at the Institute of Future Cities, The Chinese University of Hong Kong, from February 2017 to September 2017, respectively. His current research interests include noise modeling, image restoration, and diffusion models.
\end{IEEEbiography}

\begin{IEEEbiography}[{\includegraphics[width=1in,height=1.25in,clip,keepaspectratio]{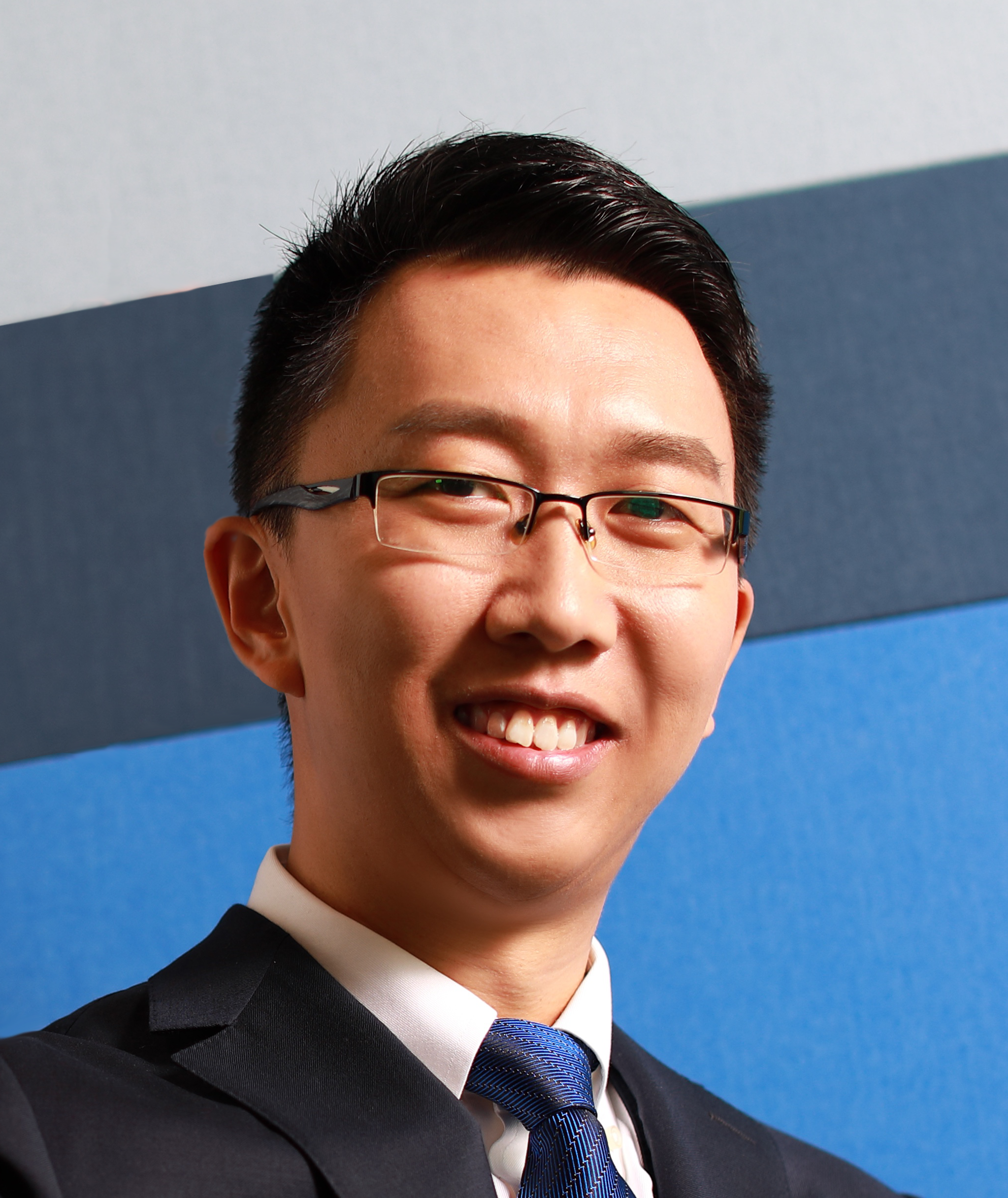}}]{Chen Change Loy}
(Senior Member, IEEE) is currently a Professor with the College of Computing and Data Science, Nanyang Technological University, Singapore. He received the PhD degree in computer science from the Queen Mary University of London, in 2010. Prior to joining NTU, he served as a research assistant professor with the Department of Information Engineering, The Chinese University of Hong Kong, from 2013 to 2018. His research interests include computer vision and deep learning with a focus on image/video restoration and enhancement, generative tasks, and representation learning. He serves as an associate editor of the IEEE Transactions on Pattern Analysis and Machine Intelligence and the International Journal of Computer Vision. He also serves/served as an Area Chair of top conferences such as ICCV, CVPR, ECCV, NeurIPS and ICLR.
\end{IEEEbiography}

\clearpage
\onecolumn
\appendix

\subsection{Mathematical Details} \label{subsec:math_supp}
\begin{itemize}[topsep=0pt,parsep=0pt,leftmargin=18pt]
    \item \textbf{Derivation of Eq.~\eqref{eq:transit_0_t}}:
    According to the transition distribution of Eq.~\eqref{eq:transit_t_t1} of our manuscript, $\bm{x}_t$ can be sampled via the following reparameterization trick:
    \begin{equation}
        \bm{x}_t = \bm{x}_{t-1} + \alpha_t \bm{e}_0 + \kappa \sqrt{\alpha_t} \bm{\xi}_t,
        \label{eq:reparametrization_xt}
    \end{equation}
    where $\bm{\xi}_t \sim \mathcal{N}(\bm{x}|\bm{0}, \bm{I})$, $\alpha_t=\eta_t-\eta_{t-1}$ for $t>1$ and $\alpha_1=\eta_1$.
    
    Applying this sampling trick recursively, we can build up the relation between $\bm{x}_t$ and $\bm{x}_0$ as follows:
    \begin{align}
        \bm{x}_t &= \bm{x}_{0} + \sum_{i=1}^t\alpha_i \bm{e}_0 + \kappa \sum_{i=1}^t \sqrt{\alpha_i} \bm{\xi}_i \notag \\
                 &= \bm{x}_{0} + \eta_t \bm{e}_0 + \kappa \sum_{i=1}^t \sqrt{\alpha_i} \bm{\xi}_i,
        \label{eq:relation_xt_x0}
    \end{align}
    where $\bm{\xi}_i \sim \mathcal{N}(\bm{x}|\bm{0}, \bm{I})$. 
    
    We can further merge $\bm{\xi}_1, \bm{\xi}_2, \cdots, \bm{\xi}_t$ and simplify Eq.~\eqref{eq:relation_xt_x0} as follows: 
    \begin{equation}
        \bm{x}_t = \bm{x}_{0} + \eta_t \bm{e}_0 + \kappa \sqrt{\eta_t} \bm{\xi_t}.
        \label{eq:reparamterization_xt_x0}
    \end{equation}
    Then the marginal distribution of Eq.~\eqref{eq:transit_0_t} in the main text is obtained based on Eq.~\eqref{eq:reparamterization_xt_x0}. 
    \item \textbf{Derivation of Eq.~\eqref{eq:poster_elbo}}: According to Bayes’s theorem, we have
    \begin{equation}
        q(\bm{x}_{t-1}|\bm{x}_t,\bm{x}_0,\bm{y}_0) \propto q(\bm{x}_t|\bm{x}_{t-1},\bm{y}_0) q(\bm{x_{t-1}}|\bm{x}_0, \bm{y}_0),
    \end{equation}
    where
    \begin{gather}
        q(\bm{x}_t|\bm{x}_{t-1},\bm{y}_0) = \mathcal{N} (\bm{x}_t;\bm{x}_{t-1}+\alpha_t \bm{e}_0, \kappa^2\alpha_t \bm{I}), \notag \\
        q(\bm{x_{t-1}}|\bm{x}_0, \bm{y}_0) = \mathcal{N} (\bm{x}_{t-1};\bm{x}_0+\eta_{t-1} \bm{e}_0, \kappa^2\eta_{t-1} \bm{I}).
    \end{gather}

    We now focus on the quadratic form in the exponent of $q(\bm{x}_{t-1}|\bm{x}_t,\bm{x}_0,\bm{y}_0)$, namely,
    \begin{align}
        &\mathrel{\phantom{=}} -\frac{(\bm{x}_t - \bm{x}_{t-1}-\alpha_t \bm{e}_0)(\bm{x}_t - \bm{x}_{t-1}-\alpha_t \bm{e}_0)^T}{2\kappa^2\alpha_t} - \frac{(\bm{x}_{t-1}-\bm{x}_0-\eta_{t-1}\bm{e}_0)(\bm{x}_{t-1}-\bm{x}_0-\eta_{t-1}\bm{e}_0)^T}{2\kappa^2\eta_{t-1}} \notag \\
        &= -\frac{1}{2}\left[\frac{1}{\kappa^2\alpha_t}+\frac{1}{\kappa^2\eta_{t-1}}\right] \bm{x}_{t-1}\bm{x}_{t-1}^T + \left[\frac{\bm{x}_t-\alpha_t \bm{e}_0}{\kappa^2\alpha_t} + \frac{\bm{x}_0+\eta_{t-1}\bm{e}_0}{\kappa^2\eta_{t-1}} \right] \bm{x}_{t-1}^T + \text{const} \notag \\
        &= - \frac{(\bm{x}_{t-1}-\bm{\mu})(\bm{x}_{t-1}-\bm{\mu})^T} {2\lambda^2} + \text{const} \label{eq:quadratic_match}
    \end{align}
    where
    \begin{equation}
        \bm{\mu}= \frac{\eta_{t-1}}{\eta_t}\bm{x}_t + \frac{\alpha_t}{\eta_t} \bm{x}_0, ~
        \lambda^2 = \kappa^2 \frac{\eta_{t-1}}{\eta_t} \alpha_t, 
    \end{equation}
    and const denotes the item that is independent of $\bm{x}_{t-1}$. This quadratic form induces the Gaussian distribution of Eq.~\eqref{eq:poster_elbo} in our manuscript.
\end{itemize}
\begin{table}[t]
    \centering
    \caption{Performance comparison of ResShift on the \textit{ImageNet-Test} dataset for image super-resolution under different configurations.}
    \label{tab:schedules}
    \small
    \vspace{-2mm}
    \begin{tabular}{@{}C{2.0cm}@{}|@{}C{2.0cm}@{}|@{}C{2.0cm}@{}|
                    @{}C{2.4cm}@{} @{}C{2.4cm}@{} @{}C{2.4cm}@{}}
         \Xhline{0.8pt}
         \multicolumn{3}{c|}{Configurations} & \multicolumn{3}{c}{Metrics} \\
         \Xhline{0.4pt}
         $T$ & $p$ & $\kappa$ & PSNR$\uparrow$ & SSIM$\uparrow$ & LPIPS$\downarrow$ \\
         \Xhline{0.4pt} 
         4    & \multirow{5}*{0.3} & \multirow{5}*{2.0} & 25.64 & 0.6903 & 0.3242  \\
         10   &                    &                    & 25.20 & 0.6828 & 0.2517  \\
         15   &                    &                    & 25.01 & 0.6769 & 0.2312  \\
         30   &                    &                    & 24.52 & 0.6585 & 0.2253  \\
         50   &                    &                    & 24.22 & 0.6483 & 0.2212  \\
        \hline \hline
        \multirow{5}*{15} & 0.3    & \multirow{5}*{2.0} & 25.01 & 0.6769 & 0.2312  \\
                          & 0.5    &                    & 25.05 & 0.6745 & 0.2387  \\
                          & 1.0    &                    & 25.12 & 0.6780 & 0.2613  \\
                          & 2.0    &                    & 25.32 & 0.6827 & 0.3050  \\
                          & 3.0    &                    & 25.39 & 0.5813 & 0.3432  \\
        \hline \hline
        \multirow{5}*{15} & \multirow{5}*{0.3} & 0.5    & 24.90 & 0.6709 & 0.2437  \\
                          &                    & 1.0    & 24.84 & 0.6699 & 0.2354  \\
                          &                    & 2.0    & 25.01 & 0.6769 & 0.2312  \\
                          &                    & 8.0    & 25.31 & 0.6858 & 0.2592  \\
                          &                    & 16.0   & 24.46 & 0.6891 & 0.2772  \\
         \Xhline{0.4pt}
    \end{tabular}   
    \vspace{-2mm}
\end{table}
\begin{figure*}[t]
    \centering
    \includegraphics[width=1.0\linewidth]{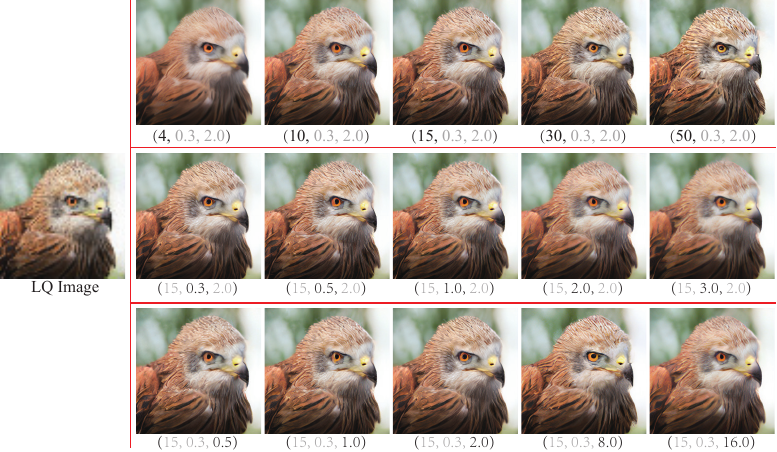}
    \caption{Qualitative comparisons of ResShift under different combinations of ($T$, $p$, $\kappa$) on the task of image super-resolution. For example, ``(15, 0.3, 2.0)'' represents the recovered result with $T=15$, $p=0.3$, and $\kappa=2.0$. Please zoom in for a better view.}
    \label{fig:ablation_schedule}
\end{figure*}

\subsection{Experimental Results on Image Super-resolution}
\subsubsection{Degradation Settings of the Synthetic Dataset}\label{subsec:degradation_app}
We synthesize the testing dataset \textit{ImageNet-Test} based on the degradation model in RealESRGAN~\cite{wang2021real} but remove the second-order operation. We observed that the low-quality (LQ) image generated by the pipeline with second-order degradation exhibited significantly more pronounced corruption than most of the real-world LQ images, we thus discarded the second-order operation to align the authentic degradation better. Next, we gave the detailed configuration of the blurring kernel, downsampling operator, and noise types.

\noindent\textbf{Blurring kernel.} The blurring kernel is randomly sampled from the isotropic Gaussian and anisotropic Gaussian kernels with a probability of [0.6, 0.4]. The window size of the kernel is set to 13. For isotropic Gaussian kernel, the kernel width is uniformly sampled from [0.2, 0.8]. For an anisotropic Gaussian kernel, the kernel widths along the $x$-axis and $y$-axis are both randomly sampled from [0.2, 0.8].

\noindent\textbf{Downampling.} We downsample the image using the ``interpolate'' function of PyTorch~\cite{paszke2019pytorch}. The interpolation mode is randomly selected from ``area'', ``bilinear'', and ``bicubic''.

\noindent\textbf{Noise.} We first add Gaussian and Poisson noise with a probability of [0.5, 0.5]. For Gaussian noise, the noise level is randomly chosen from [1,15]. For Poisson noise, we set the scale parameter in [0.05, 0.3]. Finally, the noisy image is further compressed using JPEG with a quality factor ranging in [70, 95].
\begin{figure*}[t]
    \centering
    \includegraphics[width=\linewidth]{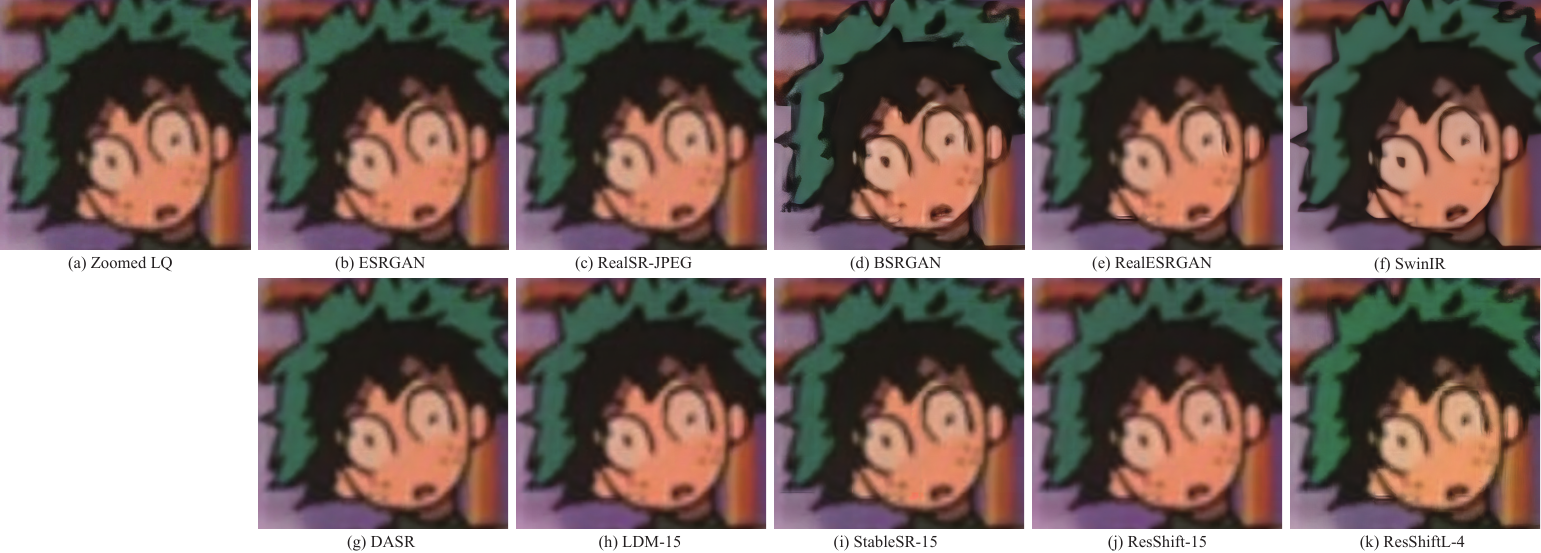}
    \caption{One typical real-world failed case in the task of image super-resolution.}
    \label{fig:failed_all}
    \vspace{-2mm}
\end{figure*}

\subsubsection{Model Analysis} \label{subsubsec:analysis_sr_app}

\vspace{2mm}\noindent\textbf{Diffusion Steps $T$ and Hyper-parameter $p$}. \zsyrevise{The proposed transition distribution in our method significantly reduces the diffusion steps $T$ in the Markov chain. The hyper-parameter $p$ allows for flexible control over the rate of residual shifting during the transition. Performance evaluations of ResShift on the test dataset of \textit{ImageNet-Test}, under various configurations of $T$ and $p$, are presented in Table~\ref{tab:schedules}. This comparison reveals that both $T$ and $p$ render an evident trade-off between the fidelity (measured by the reference metrics of PSNR and SSIM) and the perceptual quality (measured by LPIPS) of the super-resolved results. Taking the hyper-parameter $p$ as an example, an upward adjustment of its value is associated with enhancements in fidelity-oriented metrics while concurrently resulting in a deterioration in perceptual quality. Furthermore, the visual comparison in Fig.~\ref{fig:ablation_schedule} shows that a large value of $p$ will suppress the model's ability to hallucinate more image details, thereby yielding blurry outputs.}

\begin{wrapfigure}[17]{r}[0pt]{7.5cm}
    \centering
    \vspace{-4mm}
    \includegraphics[scale=0.48]{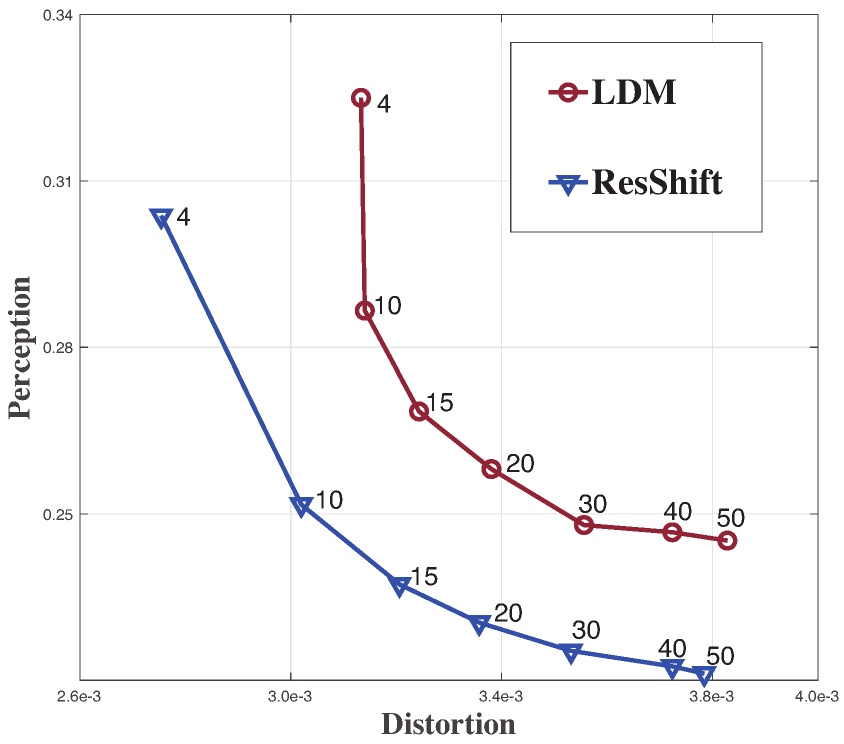}
    \vspace{-3mm}
    \caption{Perception-distortion trade-off of ResShift and LDM. The vertical and horizontal axes represent the strength of the perception and distortion, measured by LPIPS and MSE, respectively.}
   \label{fig:trade-off} 
\end{wrapfigure}
\vspace{2mm}\noindent\textbf{Perception-Distortion Trade-off.} \zsyrevise{There exists a well-known phenomenon called perception-distortion trade-off~\cite{Blau_2018_CVPR} in SR. In particular, the augmentation of the generative capability of a restoration model, such as increasing the sampling steps for a diffusion-based method or amplifying the weight of the adversarial loss for a GAN-based method, will result in a deterioration in fidelity preservation while concurrently enhancing the realism of restored images. 
In Fig.~\ref{fig:trade-off}, we plot the perception-distortion curves of ResShift and the representative baseline method LDM~\cite{rombach2022high}, wherein the perception and distortion are measured by LPIPS and mean square-error (MSE), respectively. This plot reflects the perception quality and the reconstruction fidelity of ResShift and LDM across varying numbers of diffusion steps from 4 to 50. As can be observed, the perception-distortion curve of ResShift consistently resides beneath that of the LDM, indicating its superiority in balancing perception and distortion.}

\vspace{2mm}\noindent\textbf{Hyper-parameter $\kappa$}. \zsyrevise{The hyper-parameter $\kappa$ dominates the noise strength in state $\bm{x}_t$. In Table~\ref{tab:schedules}, we report the influence of varying $\kappa$ values on the performance of ResShift. Combining this quantitative comparison with the visualization in Fig.~\ref{fig:ablation_schedule}, we can find that excessively large or small values of $\kappa$ will smooth the recovered results, regardless of their favorable metrics of PSNR and SSIM. When $\kappa$ is in the range of $[1.0,2.0]$, our method achieves the most realistic quality, as evidenced by LPIPS, which is more desirable in real applications. We thus set $\kappa$ to be $2.0$ in this work.}

\begin{table}[t]
    \centering
    \caption{\zsyrevise{Quantitative comparison of the proposed ResShiftL and LDM with various accelerated sampling algorithms on the dataset of \textit{ImageNet-Test} for image super-resolution.}}
    \label{tab:accelerat-alg}
    \small
    \vspace{-2mm}
    \begin{tabular}{@{}C{3.0cm}@{}|@{}C{2.4cm}@{}|@{}C{2.2cm}@{} @{}C{2.2cm}@{}
                    @{}C{2.2cm}@{} @{}C{2.2cm}@{}@{}C{2.2cm}@{}}
         \Xhline{0.8pt}
         Methods         & Sampler & PSNR$\uparrow$  & SSIM$\uparrow$   & LPIPS$\downarrow$   & CLIPIQA$\uparrow$ & MUSIQ$\uparrow$  \\
         \Xhline{0.4pt}
         LDM-4           & DDIM    & 24.74 & 0.6573 & 0.3452  & 0.3717  & 38.1612 \\
         LDM-4           & DPM     & 24.84 & 0.6667 & 0.2773  & 0.5027  & 46.2975 \\
         LDM-4           & PLMS    & 20.56 & 0.4432 & 0.4616  & 0.7140  & 58.7399 \\
         ResShift-4 & -  & 25.02 & 0.6833 & 0.2076  & 0.5976  & 51.9656 \\
         \Xhline{0.4pt}
    \end{tabular}   
    \vspace{-2mm}
\end{table}
\begin{figure*}[t]
    \centering
    \includegraphics[width=\linewidth]{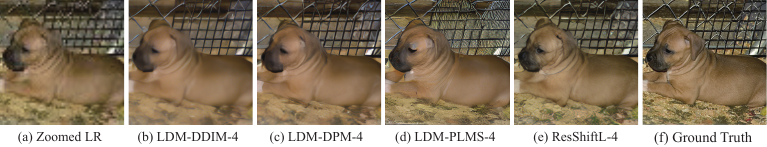}
    \vspace{-6mm}
    \caption{\zsyrevise{Qualitative comparison of the proposed ResShiftL and LDM with various accelerated sampling algorithms on the dataset of \textit{ImageNet-Test} for image super-resolution. For a fair comparison, we set the diffusion steps as 4 for LDM.}}
    \label{fig:sample_alg}
\end{figure*}
\vspace{2mm}\noindent\textbf{Comparison to LDM with More Advanced Samplers}. \zsyrevise{We conducted additional experiments to compare the proposed method with LDM accelerated by more advanced samplers, including DPM~\cite{lu2022dpmsolver} and PLMS~\cite{liu2022pseudo}. The quantitative comparisons are presented in Table~\ref{tab:accelerat-alg}, with corresponding visual results shown in Fig.~\ref{fig:sample_alg}. To ensure a comprehensive comparison, we also adopted two non-reference metrics, namely CLIPIQA~\cite{wang2022exploring} and MUSIQ~\cite{ke2021musiq}, in Table~\ref{tab:accelerat-alg}. These results clearly indicate that even with the advanced DPM algorithm, LDM~\cite{rombach2022high} still obviously underperforms compared to the proposed ResShiftL. While the use of the PLMS algorithm shows notable improvements in non-reference metrics, it compromises fidelity and introduces noticeable artifacts, as illustrated by the qualitative results in Fig.~\ref{fig:sample_alg}. Considering the high requirement on the fidelity of IR, our method proves to be more suitable for solving IR tasks.}

\subsubsection{Limitation}
Albeit its overall strong performance, the proposed method occasionally exhibits failures. One such instance is illustrated in Fig.~\ref{fig:failed_all}, where it cannot produce satisfactory results for a severely degraded comic image. It should be noted that other comparison methods also struggle to address this particular example. 
This is not an unexpected outcome as most modern image super-resolution (SR) methods are trained on synthetic datasets simulated by manually assumed degradation models~\cite{zhang2021designing,wang2021real}, which still cannot cover the full range of complicated real degradation types. Therefore, developing a more practical degradation model for SR is an essential avenue for future research.

\subsection{Experimental Results on Blind Face Restoration} \label{subsec:exp_bfr_app}
\subsubsection{Experimental Setup} \label{subsec:exp_bfr_setup} \noindent\textbf{Training Settings.} Our model was trained on the FFHQ dataset~\cite{karras2019style} that contains 70k high-quality (HQ) face images. We firstly resized the HQ images into a resolution of $512\times 512$, and then synthesized the LQ images following a typical degradation model used in recent literature~\cite{wang2021towards}:
\begin{equation}
    \bm{y} = \left\{\left[\left(\bm{x}*\bm{k}_l\right)\downarrow_s + \bm{n}_{\sigma}\right]_{\text{JPEG}_q}\right\}{\uparrow_s},
    \label{eq:degradation_synthetic}
\end{equation}
where $\bm{y}$ and $\bm{x}$ are the LQ and HQ image, $\bm{k}_l$ is the Gaussian kernel with kernel width $l$, $\bm{n}_{\sigma}$ is Gaussian noise with standard deviation $\sigma$, $*$ is 2D convolutional operator, $\downarrow_s$ and $\uparrow_s$ are the Bicubic downsampling or upsampling operators with scale $s$, and $[\cdot]_{\text{JPEG}_q}$ represents the JPEG compression process with quality factor $q$. And the hyper-parameters $l$, $s$, $\sigma$, and $q$ are uniformly sampled from $[0.1,15]$, $[0.8,32]$, $[0,20]$, and $[30,100]$ respectively. The other training configurations were kept the same as those in image super-resolution.

\vspace{2mm}\noindent\textbf{Testing Datasets}. We evaluate ResShift on one synthetic dataset and three real-world datasets. The synthetic dataset, denoted as \textit{CelebA-Test}, contains 2,000 HQ images from CelebA-HQ~\cite{karras2018progressive}, and the corresponding LQ images are synthesized via Eq.~(\ref{eq:degradation_synthetic}). As for the real-world datasets, we consider three typical ones with different degrees of degradation, namely LFW, WebPhoto~\cite{wang2021towards}, and WIDER~\cite{zhou2022towards}. LFW consists of 1711 mildly degraded face images in the wild, which contains one image for each person in LFW dataset~\cite{huang2008labeled}. WebPhoto is made up of 407 face images crawled from the internet. Some of them are old photos with severe degradation. WIDER selects 970 face images with very heavy degradation from the WIDER Face dataset~\cite{yang2016wider}, it is thus suitable to test the robustness of different methods under severe degradation.
\begin{table*}[t]
    \centering
    \caption{Quantitative comparison of different methods on \textit{CelebA-Test} dataset for blind face restoration. The results of the diffusion-based methods are denoted as ``Method-A'', where ``A'' represents the number of sampling steps. The best and second best results are highlighted in \textbf{bold} and \underline{underline}.}
    \label{tab:celeba_testing_bfr}
    \small
    \vspace{-2mm}
    \begin{tabular}{@{}C{3.6cm}@{}|
                    @{}C{2.0cm}@{} @{}C{2.0cm}@{} @{}C{2.0cm}@{} @{}C{2.0cm}@{} @{}C{2.0cm}@{} @{}C{2.2cm}@{} @{}C{2.2cm}@{}}
        \Xhline{0.8pt}
        \multirow{2}*{Methods} & \multicolumn{7}{c}{Metrics} \\
        \Xcline{2-8}{0.4pt}
            & PSNR$\uparrow$ & SSIM$\uparrow$ & LPIPS$\downarrow$ & IDS$\downarrow$  & LMD$\downarrow$ &FID-F$\downarrow$ &FID-G$\downarrow$  \\
            \Xhline{0.4pt}
            DFDNet~\cite{li2020blind}     & 22.97 & 0.631 & 0.502 & 86.32  & 20.79 & 92.22 & 77.10  \\
            PSFRGAN~\cite{chen2021progressive}    & 22.58 & 0.628 & 0.411 & 70.32  & 7.19 & 65.65 & 62.44 \\
            GFPGAN~\cite{wang2021towards} & 22.06 & 0.629 & 0.413 & 68.78  & 8.64 & \textbf{49.15} & 56.13  \\
            RestoreFormer~\cite{wang2022restoreformer}    & 22.55 & 0.598 & 0.423 & 65.93  & 8.20 & 50.76 & 53.25 \\
            VQFR~\cite{gu2022vqfr}   & 21.80 & 0.579 & 0.424 & 67.62  & 8.46 & \underline{49.62} & 57.12 \\
            CodeFormer~\cite{zhou2022towards}   & \underline{23.58} & 0.661 & \underline{0.324} & \textbf{59.14}  & \textbf{5.04} & 64.25 & 26.65  \\
            DifFace-100~\cite{yue2022difface}    & \textbf{24.24} & \textbf{0.702} & 0.334  & 61.25  & 5.13 & 52.34 & \underline{22.84} \\
            ResShift-4  & 23.41      & \underline{0.671} & \textbf{0.309} & \underline{59.70}  & \underline{5.05}  & 52.07 & \textbf{17.84} \\   
       \Xhline{0.8pt}
    \end{tabular} 
    \vspace{-2mm}
\end{table*}
\begin{table*}[t]
    \centering
    \caption{Quantitative comparison of different methods on three real-world datasets for blind face restoration. The results of the diffusion-based methods are denoted as ``Method-A'', where ``A'' represents the number of sampling steps. The best and second best results are highlighted in \textbf{bold} and \underline{underline}.}
    \label{tab:real_testing_bfr}
    \small
    \vspace{-2mm}
    \begin{tabular}{@{}C{4.0cm}@{}|
                    @{}C{2.3cm}@{} @{}C{2.3cm}@{} | @{}C{2.3cm}@{}
                    @{}C{2.3cm}@{} | @{}C{2.3cm}@{} @{}C{2.3cm}@{}}
        \Xhline{0.8pt}
        \multirow{3}*{Methods} & \multicolumn{6}{c}{Datasets} \\
        \Xcline{2-7}{0.4pt}
        & \multicolumn{2}{c|}{LFW} &\multicolumn{2}{c|}{WebPhoto} &\multicolumn{2}{c}{WIDER} \\
        \Xcline{2-7}{0.4pt}
            & FID-F$\downarrow$ & MUSIQ$\uparrow$ & FID-F$\downarrow$ & MUSIQ$\uparrow$   & FID-F$\downarrow$ & MUSIQ$\uparrow$ \\
            \Xhline{0.4pt}
            DFDNet~\cite{li2020blind}     & 59.81 & 73.11 & 92.39 & 69.03  & 57.85 & 63.21  \\
            PSFRGAN~\cite{chen2021progressive}    & 49.65 & 73.60 & 85.03 & 71.67  & 49.85 & 71.51  \\
            GFPGAN~\cite{wang2021towards} & 50.02 & 73.57 & 87.57 & \underline{72.08}  & 39.46 & 72.82   \\
            RestoreFormer~\cite{wang2022restoreformer}    & 48.50 & 73.70 & 78.16 & 69.84  & 49.85 & 67.84  \\
            VQFR~\cite{gu2022vqfr}   & \textbf{44.14} & \underline{74.02} & \underline{75.38} & 72.00  & 50.79 & \textbf{74.74} \\
            CodeFormer~\cite{zhou2022towards}   & 52.43 & \textbf{75.49} & 83.27 & \textbf{73.99}  & 38.86 & \underline{73.40}   \\
            DifFace-100~\cite{yue2022difface}    & \underline{45.64} & 70.39 & 89.99  & 66.29  & \underline{38.40} & 65.99  \\
            ResShift-4  & 52.40    & 70.68 & \textbf{74.80} & 70.90  & \textbf{38.12}  & 71.07  \\   
       \Xhline{0.8pt}
    \end{tabular} 
    \vspace{-2mm}
\end{table*}

\vspace{2mm}\noindent\textbf{Compared Methods.} We compare ResShift with seven recent BFR methods, including DFDNet~\cite{li2020blind}, PSFRGAN~\cite{chen2021progressive}, GFPGAN~\cite{wang2021towards}, RestoreFormer~\cite{wang2022restoreformer}, VQFR~\cite{gu2022vqfr}, CodeFormer~\cite{zhou2022towards}, and DifFace~\cite{yue2022difface}.

\vspace{2mm}\noindent\textbf{Evaluation Metrics.} To comprehensively assess various methods, this study adopts six quantitative metrics following the setting of VQFR~\cite{gu2022vqfr}, namely PSNR, SSIM~\cite{wang2004image}, LPIPS~\cite{zhang2018unreasonable}, identity score (IDS), landmark distance (LMD), and FID~\cite{heusel2017gans}. Note that IDS, also referred to as "Deg" in certain literature~\cite{gu2022vqfr}, and LMD both serve as quantifiers for the identity between the restored images and their ground truths. IDS gauges the embedding angle of ArcFace~\cite{deng2019arcface}, while LMD calculates the landmark distance using $L_2$ norm between pairs of images. FID quantifies the KL divergence between the feature distributions, assumed as Gaussian distribution, of the restored images and a high-quality reference dataset. For the reference dataset, we employ both the ground truth images and the FFHQ~\cite{karras2018progressive} dataset. The corresponding results computed under these two settings are denoted as ``FID-G'' and ``FID-F'' for clarity. On the real-world datasets, we mainly adopt two no-reference metrics, namely FID-F and MUSIQ~\cite{ke2021musiq}, since the underlying ground truth images are unavailable.

\subsubsection{Evaluation on Synthetic Dataset}
We present the comparative results on \textit{CelebA-Test} in Table~\ref{tab:celeba_testing_bfr}. The proposed ResShift demonstrates superior performance, particularly in terms of LPIPS and FID-G, indicating the heightened alignment of its restored results with the perceptual system of humans. Regarding the identity-related metrics, namely LMD and IDS, our method attains the second-best rankings, substantiating its powerful capability for identity preservation. Furthermore, our method exhibits, at a minimum, comparable performance to recent state-of-the-art (SotA) techniques across other evaluated metrics. In summary, our proposed method manifests commendable and consistent proficiency in blind face restoration. 

\begin{figure*}[t]
    \centering
    \includegraphics[width=1.0\linewidth]{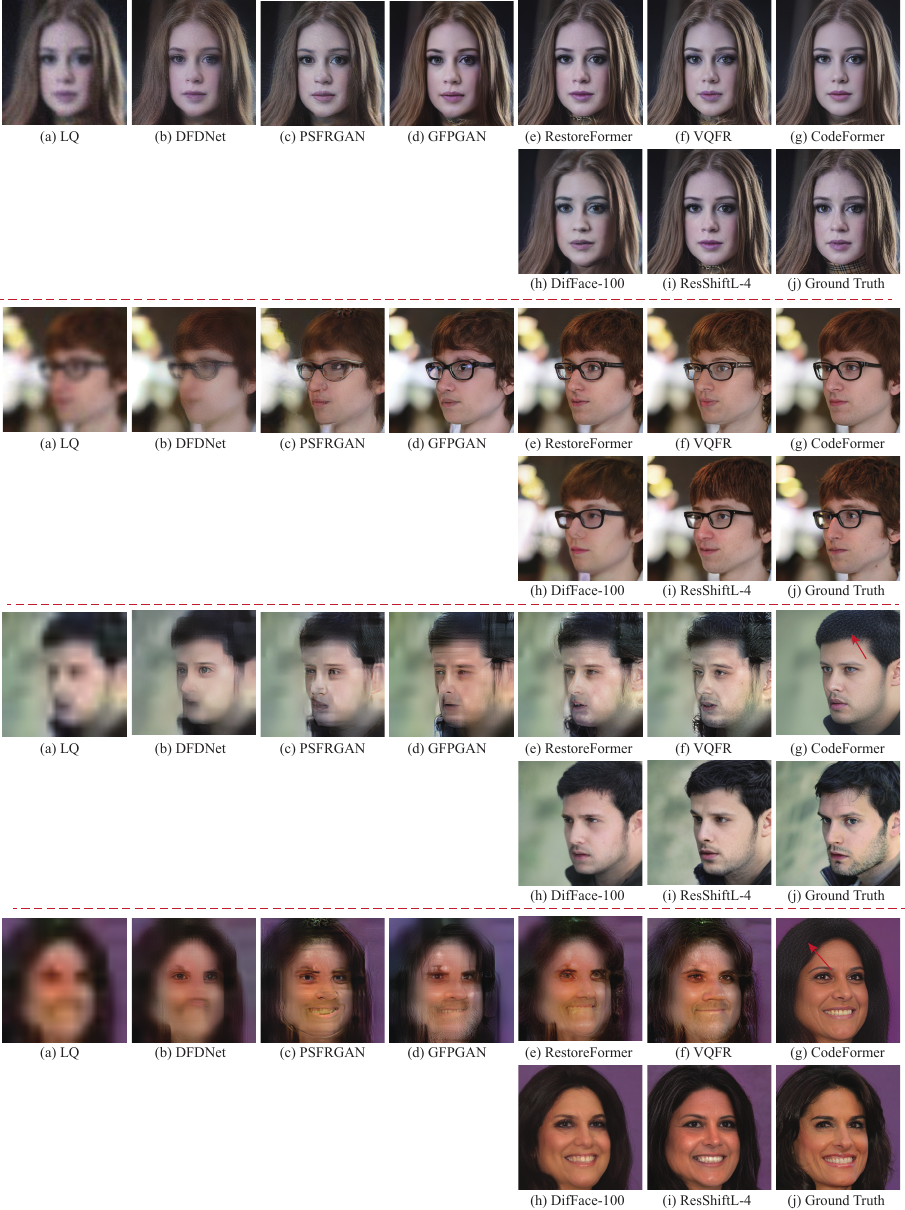}
    \vspace{-6mm}
    \caption{Qualitative results of different methods on the synthetic \textit{CelebA-Test} dataset for blind face restoration. We annotate the diffusion-based methods with the format of ``Method-A'', where ``A'' represents the number of sampling steps. Please zoom in for a better view.}
    \label{fig:syn_celeba_bfr}
\end{figure*}

For visualization, four typical examples of the \textit{CelebA-Test} are displayed in Fig.~\ref{fig:syn_celeba_bfr}. In the first and second examples with mild degradation, most of the comparison methods can restore a realistic-looking image. When confronted with more severe degradation as shown in the third and fourth examples, only CodeFormer~\cite{zhou2022towards}, DifFace~\cite{yue2022difface}, and ResShift can handle such cases, yielding satisfactory facial images. However, the results of CodeFormer still contain some slight artifacts in specific areas, such as hair, as highlighted by red arrows in Fig.~\ref{fig:syn_celeba_bfr}). As for DifFace, it needs 100 sampling steps, largely limiting its efficiency. In contrast, the proposed ResShift not only requires much fewer diffusion steps, i.e., 4 steps, but also performs more stably under this challenging degradation setting. 

\subsubsection{Evaluation on Real-world Dataset}\label{sec:exp_real_bfr}
\begin{figure*}[t]
    \centering
    \includegraphics[width=\linewidth]{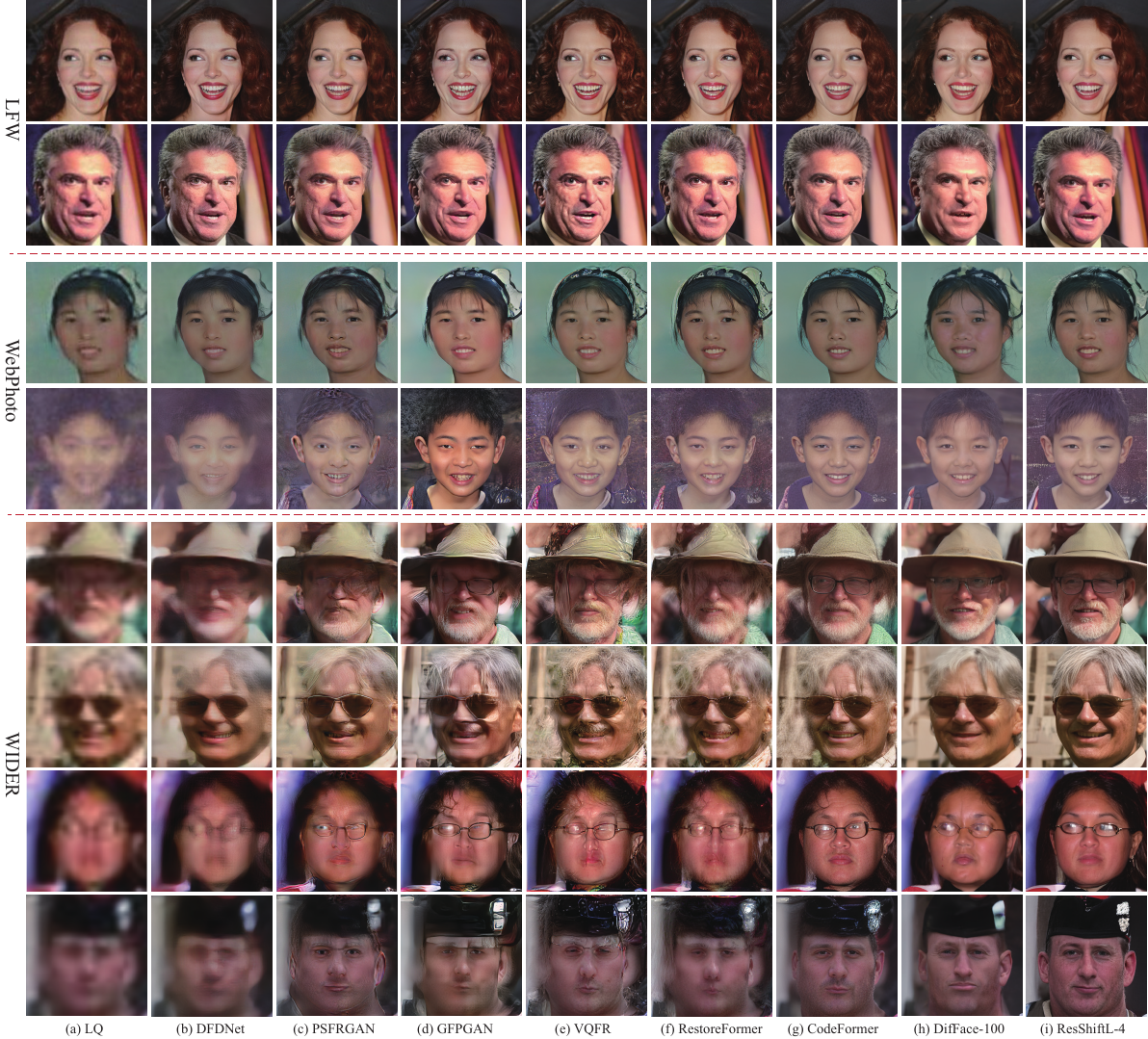}
    \vspace{-6mm}
    \caption{Qualitative results of different methods on three real-world datasets for blind face restoration. We annotate the diffusion-based methods with the format of ``Method-A'', where ``A'' represents the number of sampling steps. Please zoom in for a better view.}
    \label{fig:real_face_bfr}
\end{figure*}

The comparative results on three real-world datasets are summarized in Table~\ref{tab:real_testing_bfr}. We can observe that ResShift surpasses its counterparts with regard to the metric of FID-F, while maintaining comparability with recent SotA methodologies in terms of MUSIQ. To supplement the analysis, we show several typical examples of these datasets in Fig.~\ref{fig:real_face_bfr}. It is observed that all the comparison approaches perform well on the dataset LFW with slight degradation. However, ResShift provides significantly better results on the other two datasets where the LQ images are severely degraded. This stable performance of ResShift is consistent with the evaluation metric of FID-F, mainly owing to the powerful capability of the designed diffusion model. 

\begin{table}[t]
    \centering
    \caption{\zsyrevise{Quantitative comparison of different methods on the testing dataset of GoPro for image deblurring. The results of the diffusion-based methods are denoted as ``Method-A'', where ``A'' represents the number of sampling steps. The best and second best results are highlighted in \textbf{bold} and \underline{underline}.}}
    \label{tab:deblur_gopro}
    \small
    \vspace{-2mm}
    \begin{tabular}{@{}C{4.0cm}@{}|
                    @{}C{2.5cm}@{} @{}C{2.5cm}@{} @{}C{2.5cm}@{}
                    @{}C{2.5cm}@{} @{}C{2.5cm}@{}}
        \Xhline{0.8pt}
        \multirow{2}*{Methods} & \multicolumn{5}{c}{Metrics} \\
        \Xcline{2-6}{0.4pt}
            & PSNR$\uparrow$ & SSIM$\uparrow$ & LPIPS $\downarrow$ & FID$\downarrow$   & NIQE $\downarrow$ \\
            \Xhline{0.4pt}
            DeblurGAN-v2~\cite{kupyn2019deblur}  & 29.08 & 0.8766 & 0.1173 & \underline{14.33}  & \textbf{4.940}   \\
            MIMO-UNet+~\cite{cho2021rethinking}  & 32.44 & 0.9333 & 0.0905 & 19.96  & 5.563  \\
            MPRNet~\cite{zamir2021multi}         & 32.66 & 0.9363 & 0.0886 & 22.00  & 5.653   \\
            Uformer~\cite{wang2022uformer}       & \underline{33.05} & \underline{0.9418} & 0.0868 & 22.05  & 5.668   \\
            Restormer~\cite{zamir2022restormer}  & 32.92 & 0.9399 & 0.0841 & 21.21  & 5.683 \\
            DiffIR-4~\cite{Xia_2023_ICCV}        & \textbf{33.31} & \textbf{0.9446} & \underline{0.0787} & 20.93  & 5.674   \\
            ResShiftL-4                 & 29.47 & 0.8856 & \textbf{0.0720} & \textbf{9.39}   & \underline{5.194}  \\   
       \Xhline{0.8pt}
    \end{tabular} 
    \vspace{-2mm}
\end{table}

\subsection{Experimental Results on Image Deblurring}
\subsubsection{Experimental Setup} \label{subsec:exp_deblur_setup} \zsyrevise{We train ResShiftL on the GoPro~\cite{nah2017deep} dataset and evaluate its performance on the testing dataset of GoPro following recent work~\cite{zhang2024unified}. For a comprehensive evaluation, we employed both distortion metrics, including PSNR and SSIM~\cite{wang2004image}, as well as perceptual metrics, namely LPIPS~\cite{zhang2018unreasonable}, FID~\cite{heusel2017gans}, and NIQE~\cite{mittal2012making}. Note that the FID score was computed at the patch level by extracting non-overlapping patches of size $256\times 240$ from each $1280\times 720$ source image, as recommended by~\cite{whang2022deblurring} to obtain a stable evaluation. We compared our approach against six SotA methods: DeblurGAN-v2~\cite{kupyn2019deblur}, MIMO-Unet+~\cite{cho2021rethinking}, MPRNet~\cite{zamir2021multi}, Uformer~\cite{wang2022uformer}, Restormer~\cite{zamir2022restormer}, and DiffIR~\cite{Xia_2023_ICCV}.}

\begin{figure*}[t]
    \centering
    \includegraphics[width=1.0\linewidth]{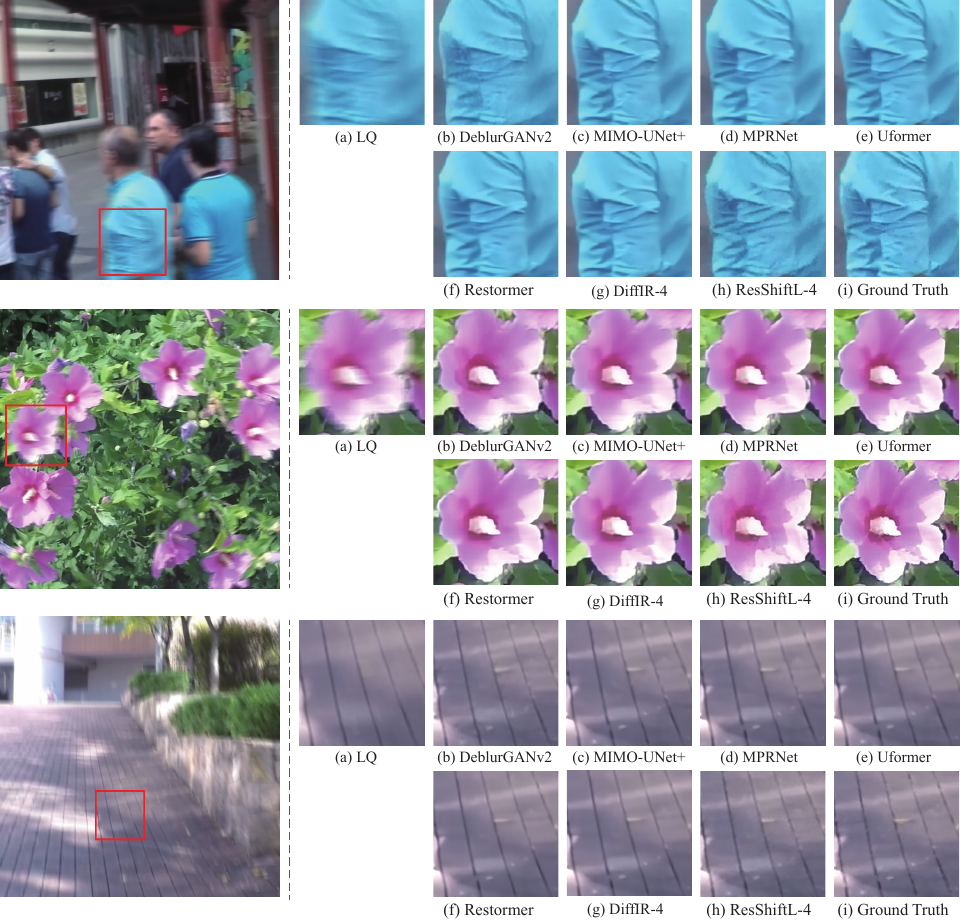}
    \vspace{-6mm}
    \caption{\zsyrevise{Qualitative results of different methods on the GoPro testing dataset for image deblurring. We annotate the diffusion-based methods with the format of ``Method-A'', where ``A'' represents the number of sampling steps. Please zoom in for a better view.}}
    \label{fig:gopro3deblur}
\end{figure*}

\subsubsection{Experimental Results} \label{subsec:exp_deblur_setup}
\zsyrevise{Table~\ref{tab:deblur_gopro} presents a comparative analysis of various methods evaluated on the GoPro~\cite{nah2017deep} testing dataset. The results indicate that the proposed ResShiftL demonstrates superior performance with respect to perceptual metrics, in particular of LPIPS and FID. This suggests that ResShiftL aligns more closely with human visual perception. Additionally, the visual evidence provided in Fig.~\ref{fig:gopro3deblur} further proves the perceptual advantages of our approach. However, in terms of distortion metrics, such as PSNR and SSIM, our method performs less favorably compared to existing methods. This is mainly because ResShiftL is implemented in the latent space of VQGAN, which is compressed by a factor of $8$. The transformation between the pixel space and latent space inevitably results in some information loss, thus limiting the performance of our method regarding distortion metrics.}

\begin{figure*}[t]
    \centering
    \includegraphics[width=\linewidth]{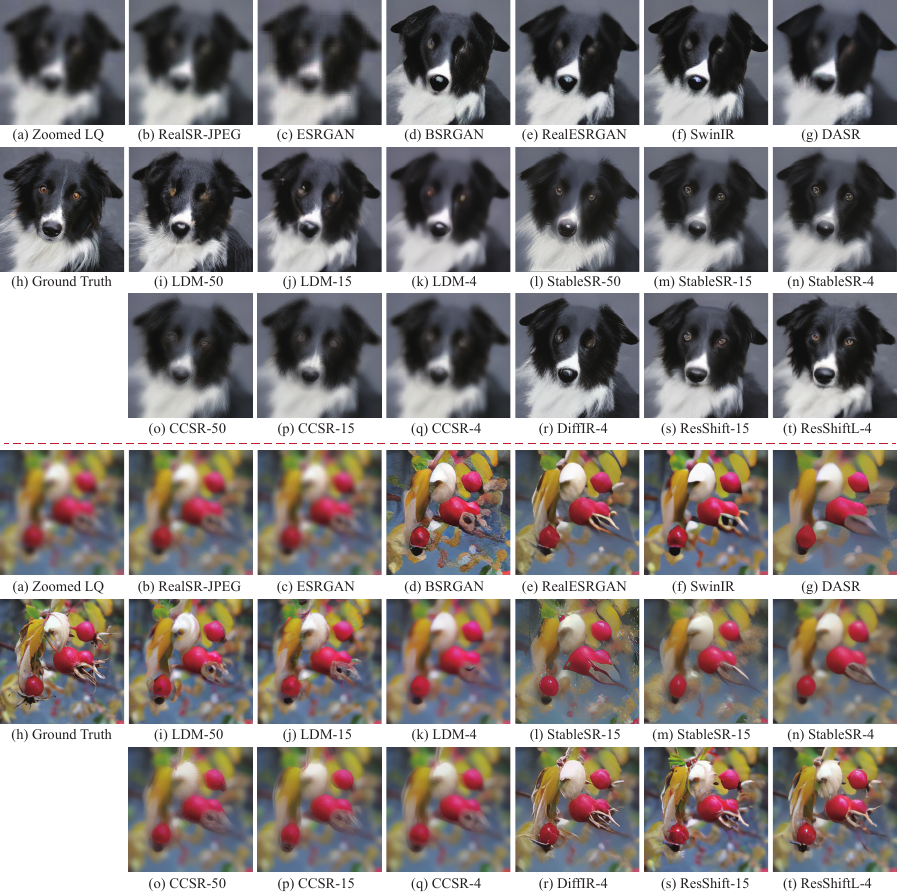}
    \vspace{-6mm}
    \caption{Qualitative results of different methods on the synthetic \textit{ImageNet-Test} dataset for image super-resolution. Please zoom in for a better view.}
    \label{fig:syn_imagenet_app}
\end{figure*}
\begin{figure*}[t]
    \centering
    \includegraphics[width=\linewidth]{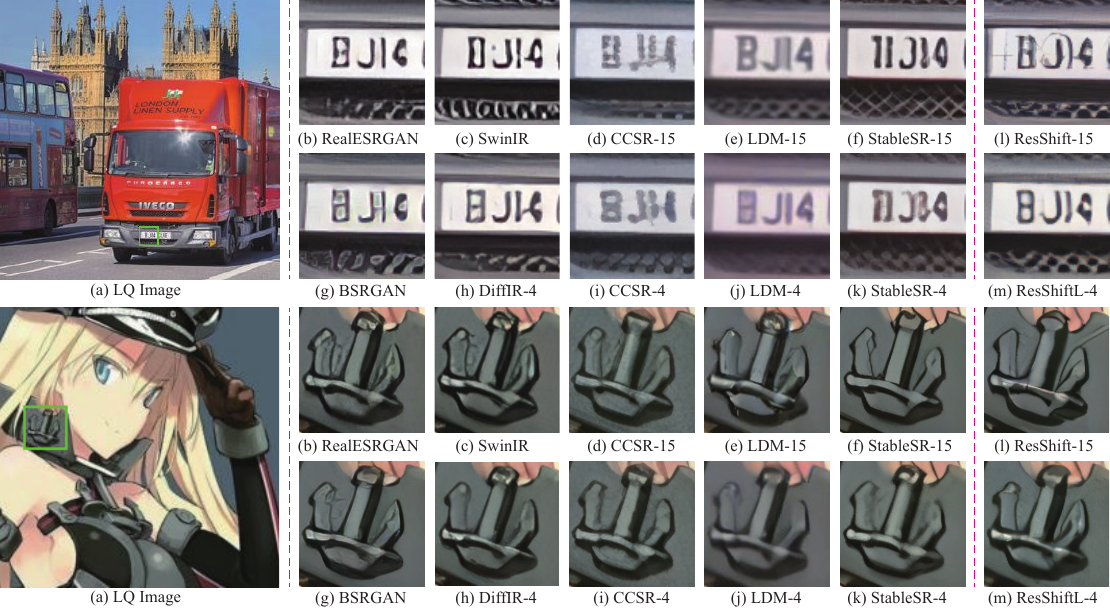}
    \vspace{-6mm}
    \caption{Qualitative comparisons on two real-world examples from \textit{RealSet80}. Please zoom in for a better view.}
    \label{fig:real_data_app}
\end{figure*}
\begin{figure*}[t]
    \centering
    \includegraphics[width=\linewidth]{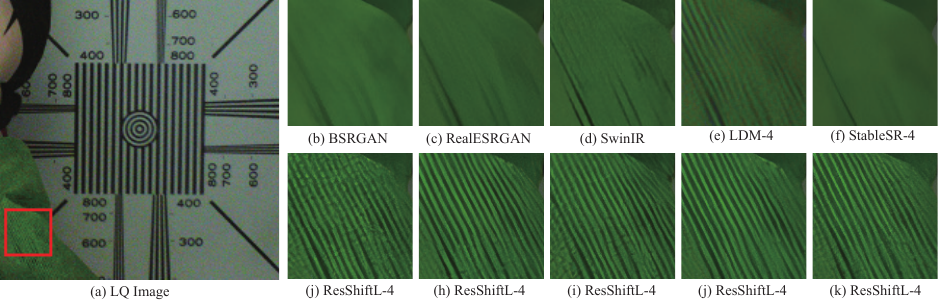}
    \vspace{-7mm}
    \caption{\zsyrevise{Visual analysis of the sampling randomness. (a) LQ image, (b)-(f) restored images by recent state-of-the-art methods, (j)-(k) restored results by our proposed method under different random seeds.}}
    \label{fig:randomness-sr}
    \vspace{-2mm}
\end{figure*}
\begin{figure*}[t]
    \centering
    \includegraphics[width=\linewidth]{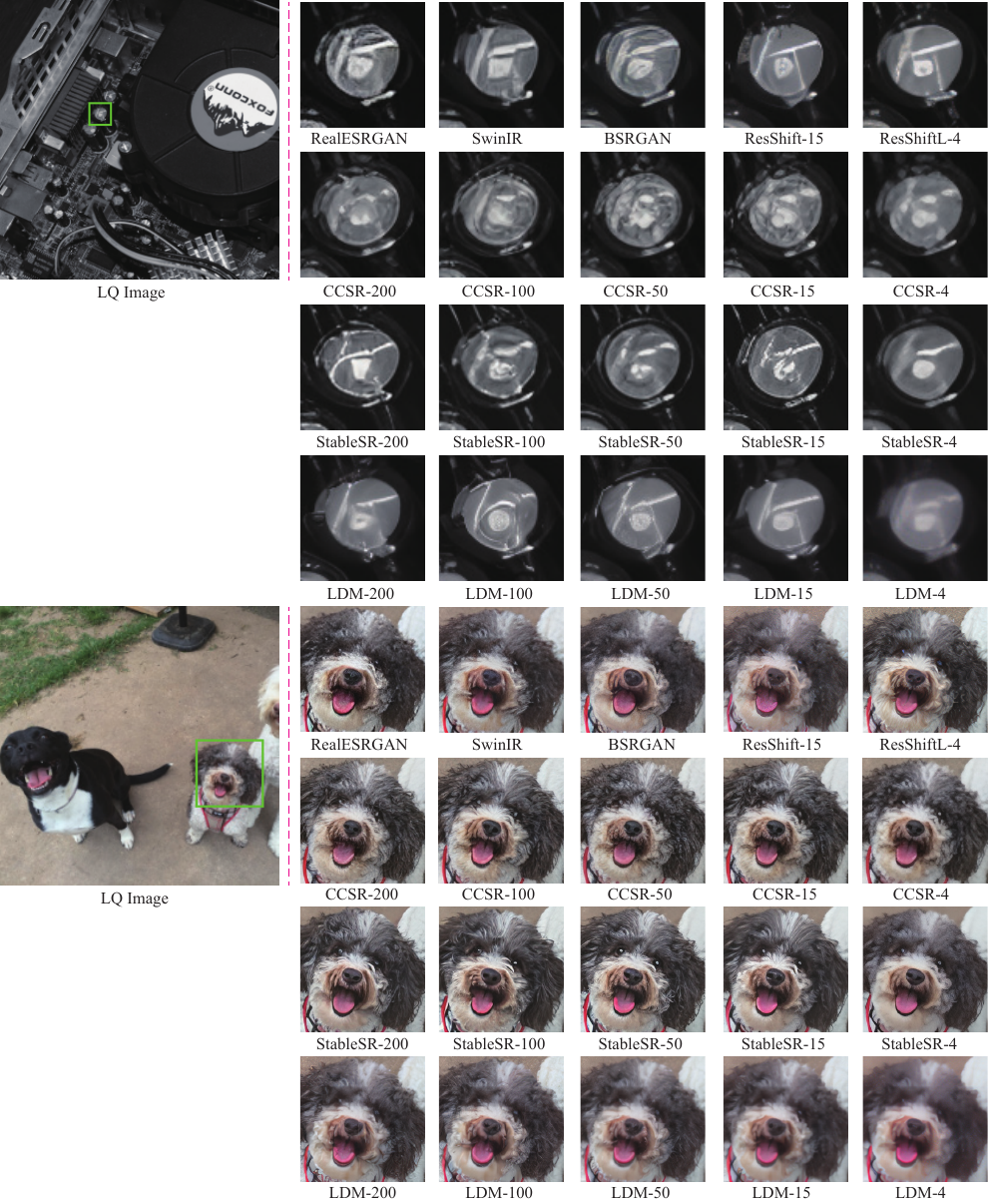}
    \vspace{-6mm}
    \caption{\zsyrevise{Qualitative comparisons on two real-world examples from \textit{RealSet80}. For the diffusion-based methods, we display the results with different sampling steps, ranging from 4 to 200. Please zoom in for a better view.}}
    \label{fig:real_data_multistep}
\end{figure*}
\begin{figure*}[t]
    \centering
    \includegraphics[width=0.82\linewidth]{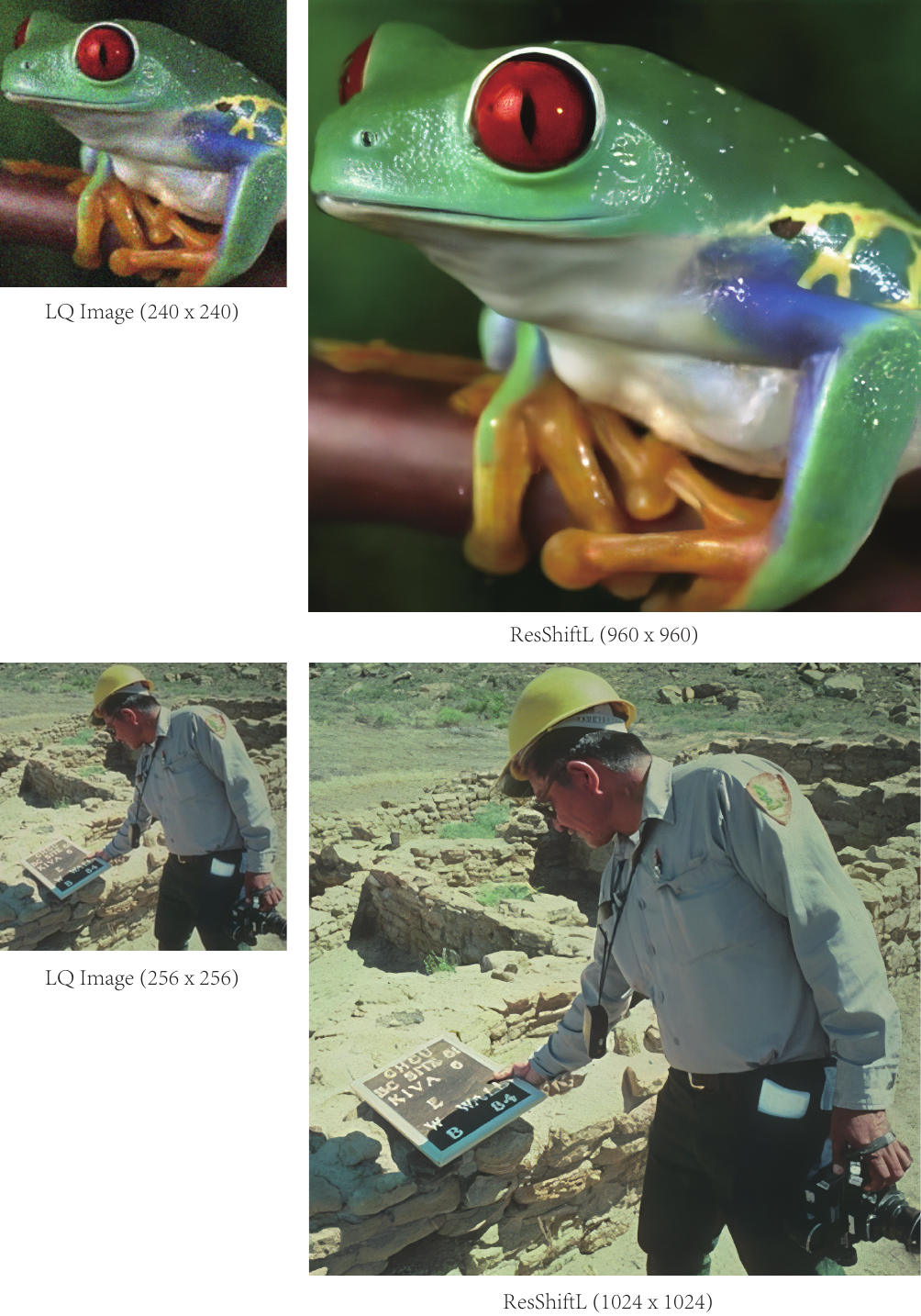}
    \vspace{-2mm}
    \caption{\zsyrevise{Super-resolution results of the proposed ResShiftL on two real-world examples with heavy degradation from \textit{RealSet80}. Top row: x4 super-resolution from $240\times 240$ to $960\times 960$. Bottom row: x4 super-resolution from $256\times 256$ to $1024\times 1024$. Please zoom in for a better view.}}
    \label{fig:heavy-deg-realset80}
\end{figure*}
\begin{figure*}[t]
    \centering
    \includegraphics[width=0.82\linewidth]{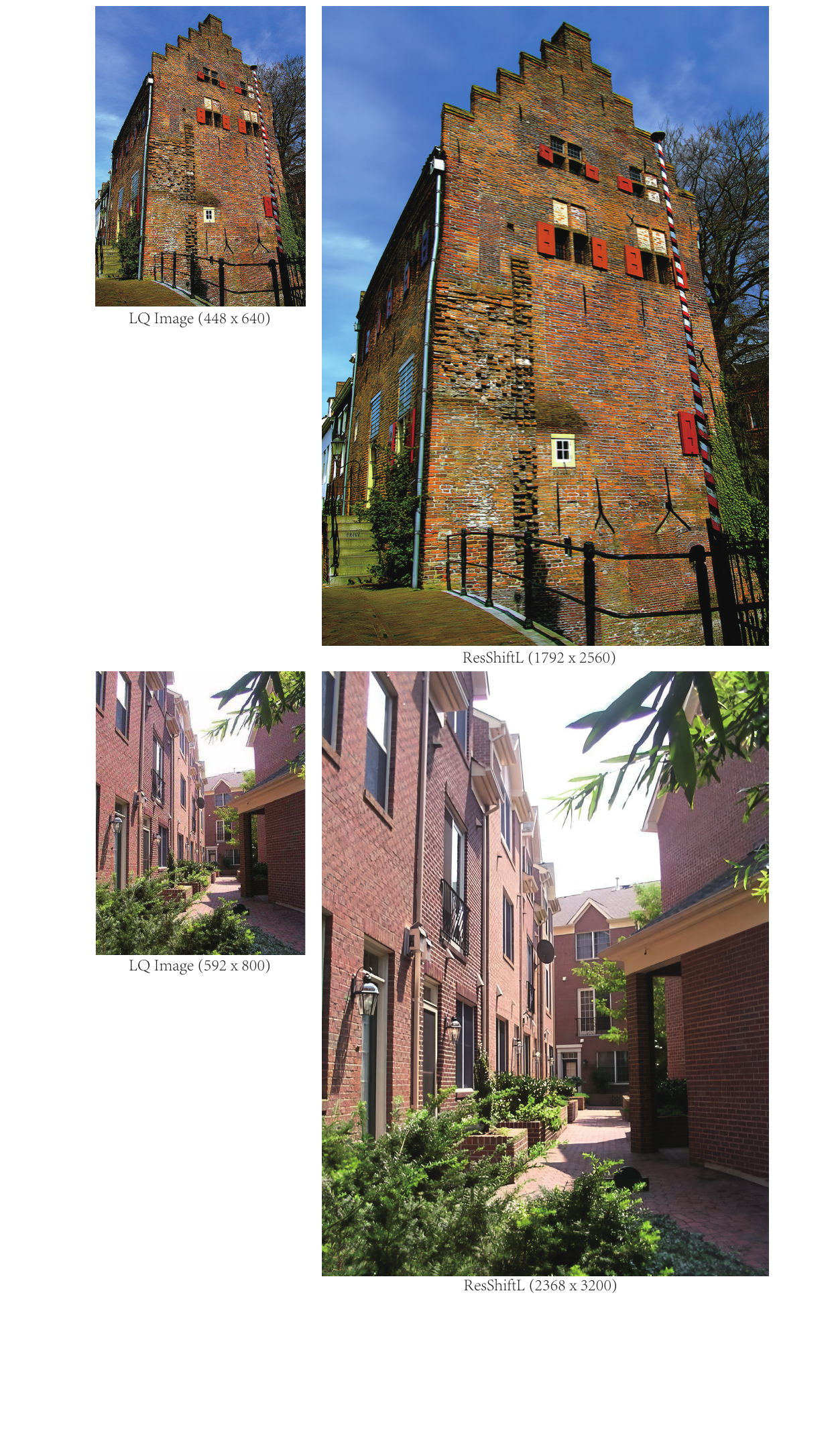}
    \vspace{-30mm}
    \caption{\zsyrevise{Super-resolution results of the proposed ResShiftL on two real-world examples with slight degradation from \textit{RealSet80}. Top row: x4 super-resolution from $448\times 640$ to $1792\times 2560$. Bottom row: x4 super-resolution from $592\times 800$ to $2368\times 3200$. Please zoom in for a better view.}}
    \label{fig:sligt-deg-realset80}
\end{figure*}
\begin{figure*}[t]
    \centering
    \includegraphics[width=0.90\linewidth]{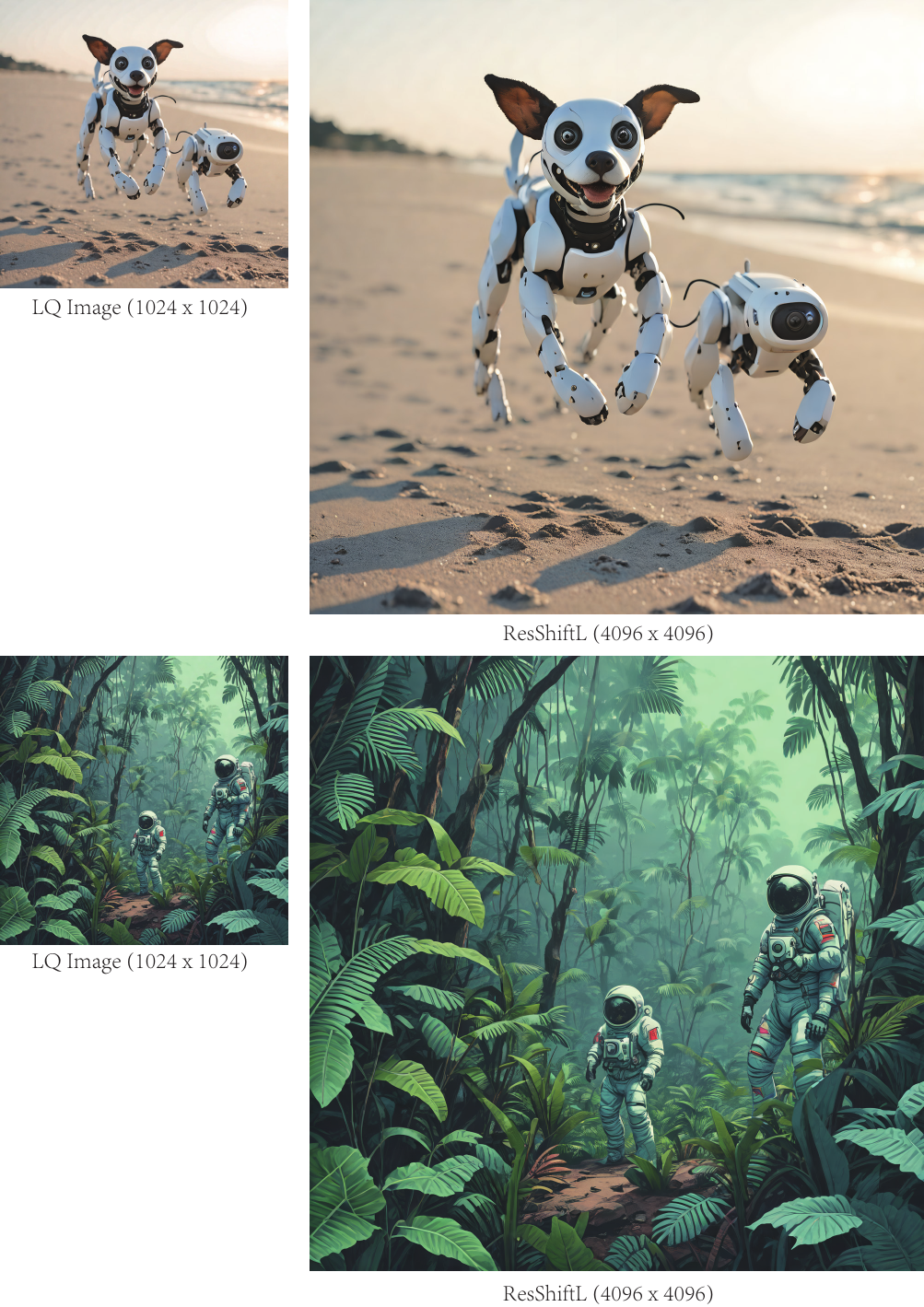}
    \vspace{-2mm}
    \caption{\zsyrevise{Super-resolution results (x4, $1024\times1024 \rightarrow 4096\times 4096$) of the proposed ResShiftL on two synthesized examples by SDXL-Turbo. Please zoom in for a better view.}}
    \label{fig:sligt-deg-sdxlturbo}
\end{figure*}
\begin{figure*}[t]
    \centering
    \includegraphics[width=\linewidth]{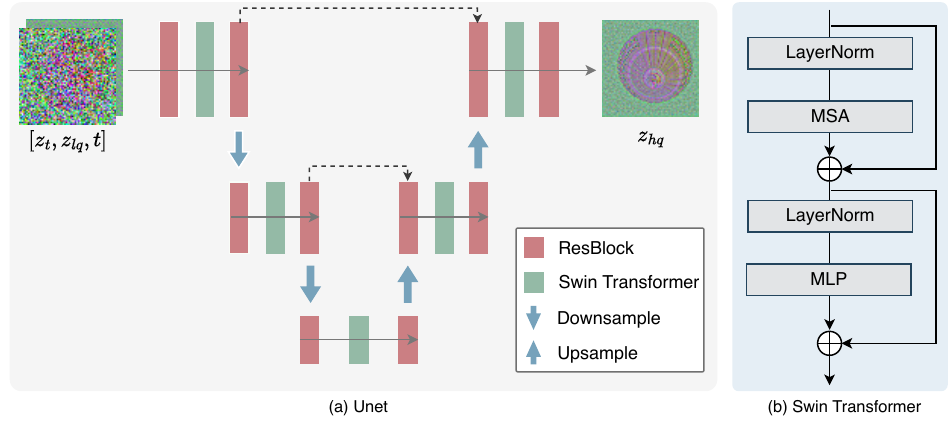}
    \vspace{-10mm}
    \caption{Illustration of the network architecture of our method. It is modified from the widely-used diffusion Unet. To better handle the images with various resolutions, we introduce several Swin Transformer blocks, each consisting of LayerNorm, Multi-head Self-Attention (MSA), and MLP.}
    \label{fig:network}
\end{figure*}

\end{document}